\documentclass{article}

\PassOptionsToPackage{usenames, dvipsnames}{xcolor}

\usepackage{microtype}
\usepackage{graphicx}
\usepackage{booktabs} %
\usepackage{hyperref}

\usepackage{tikz}

\newif\ifdraft
\drafttrue
\usepackage{rotating}
\usepackage{float}
\usepackage{setspace}
\usepackage[bottom]{footmisc}
\usepackage{amsthm}
\usepackage{amssymb}
\usepackage{amsmath}
\usepackage{paralist} %
\usepackage{pifont}
\usepackage{mathtools}
\usepackage{multirow}
\usepackage{booktabs}
\usepackage{multirow}
\usepackage{style_ltx}
\usepackage{xspace}
\usepackage{algorithm}
\usepackage{fullpage}
\usepackage{array}
\usepackage{wrapfig}
\usepackage[labelformat=simple]{subcaption}

\usepackage[utf8]{inputenc} %
\usepackage{hyperref}       %
\usepackage{url}            %
\usepackage{amsfonts}       %
\usepackage{nicefrac}       %
\usepackage[inline]{enumitem}
\usepackage[compact]{titlesec}
\newcommand{\cmark}{\ding{51}}%
\newcommand{\xmark}{\ding{55}}%
\usepackage{authblk}
\usepackage[acronym,nowarn,section,nogroupskip,nonumberlist]{glossaries}
\usepackage{cleveref}
\usepackage[dvipsnames]{xcolor}
\usepackage{algpseudocode}

\title{On Calibration of Object Detectors: Pitfalls, Evaluation and Baselines}

\author[]{Selim Kuzucu*}
\author[]{Kemal Oksuz*}
\author[]{Jonathan Sadeghi}
\author[]{Puneet K. Dokania}
\affil[]{Five AI Ltd., United Kingdom \protect\\
\{selim.kuzucu2, kemal.oksuz, jonathan.sadeghi, puneet.dokania\}@five.ai}

\date{}

\glsdisablehyper{}
\newacronym{ID}{ID}{in-distribution}
\newacronym{SAOD}{SAOD}{Self-aware Object Detection}
\newacronym{SOTA}{SOTA}{state-of-the-art}
\newacronym{TS}{TS}{Temperature Scaling}
\newacronym{AP}{AP}{Average Precision}
\newacronym{LaECE}{LaECE}{Localisation-aware Expected Calibration Error}
\newacronym{LaACE}{LaACE}{Localisation-aware Adaptive Calibration Error}
\newacronym{CE}{CE}{Calibration Error}
\newacronym{IR}{IR}{Isotonic Regression}
\newacronym{LR}{LR}{Linear Regression}
\newacronym{PS}{PS}{Platt Scaling}
\newacronym{ELCE}{ELCE}{Expected Localization Calibration Error}
\newacronym{ECE}{ECE}{Expected Calibration Error}
\newacronym{DECE}{D-ECE}{Detection Expected Calibration Error}
\newacronym{AUC}{AUC}{area-under-curve}
\newacronym{TN}{TN}{true-negative}
\newacronym{TP}{TP}{true-positive}
\newacronym{FP}{FP}{false-positive}
\newacronym{FN}{FN}{false-negative}
\newacronym{IoU}{IoU}{Intersection-over-Union}
\newacronym{LRP}{LRP}{Localisation-Recall-Precision Error}
\newacronym{AV}{AV}{Autonomous Vehicles}
\newacronym{NMS}{NMS}{Non-maximum Suppression}
\newacronym{OKS}{OKS}{Object Keypoint Similarity}
\newacronym{NLL}{NLL}{Negative Log-Likelihood}
\newacronym{PR}{PR}{Precision Recall}

\newcommand{\blockcomment}[1]{}
\definecolor{forestgreen}{rgb}{0.13, 0.55, 0.13}
\newcommand{\imp}[1]{\textcolor{forestgreen}{\small{(+#1)}}}
\newcommand{\nimp}[1]{\textcolor{red}{\small{(-#1)}}}

\newcommand{\testdata}{\mathcal{D}_{\mathrm{Test}}}
\newcommand{\traindata}{\mathcal{D}_{\mathrm{Train}}}
\newcommand{\valdata}{\mathcal{D}_{\mathrm{Val}}}

\footnotetext{Equal contributions. SK contributed during his internship at Five AI Oxford team.}

\definecolor{DarkGreen}{rgb}{0, 0.4, 0}

\graphicspath{{./figs/},{./images_adv/}}

\usetikzlibrary{arrows,decorations.pathmorphing,backgrounds,fit,positioning,shapes.symbols,chains}
\tikzset{ node distance = 1cm, auto,font=\footnotesize,
tensors/.style={circle, rounded corners, draw=black, fill=black!10, inner sep=0.5pt, text width=1cm, text badly centered, minimum height=1.2cm,, font=\bfseries\footnotesize\sffamily} ,
temp_tensors/.style={circle, rounded corners, dashed, draw=black, fill=black!5, inner sep=0.5pt, text width=1cm, text badly centered, minimum height=1.2cm,, font=\bfseries\footnotesize\sffamily} ,
parameters/.style={align=center, text width=2cm, font=\bfseries\footnotesize\sffamily}}

\newcolumntype{L}[1]{>{\raggedright\let\newline\\\arraybackslash\hspace{0pt}}m{#1}}
\newcolumntype{C}[1]{>{\centering\let\newline\\\arraybackslash\hspace{0pt}}m{#1}}
\newcolumntype{R}[1]{>{\raggedleft\let\newline\\\arraybackslash\hspace{0pt}}m{#1}}

\begin{document}
\maketitle
\begin{abstract}
Reliable usage of object detectors require them to be calibrated---a crucial problem that requires careful attention. 
Recent approaches towards this involve (1) designing new loss functions to obtain calibrated detectors by training them from scratch, and (2) post-hoc Temperature Scaling (TS) that learns to scale the likelihood of a trained detector to output calibrated predictions. These approaches are then evaluated based on a combination of Detection Expected Calibration Error (D-ECE) and Average Precision.
In this work, via extensive analysis and insights, we highlight that these recent evaluation frameworks, evaluation metrics, and the use of TS have notable drawbacks leading to incorrect conclusions.
As a step towards fixing these issues, we propose a principled evaluation framework to jointly measure calibration and accuracy of object detectors.
We also tailor efficient and easy-to-use post-hoc calibration approaches such as Platt Scaling and Isotonic Regression specifically for object detection task.
Contrary to the common notion, our experiments show that once designed and evaluated properly, post-hoc calibrators, which are extremely cheap to build and use, are much more powerful and effective than the recent train-time calibration methods.
To illustrate, D-DETR with our \textit{post-hoc} Isotonic Regression calibrator outperforms the recent \textit{train-time} state-of-the-art calibration method Cal-DETR~\cite{munir2023caldetr} by more than $7$ D-ECE on the COCO dataset. Additionally, we propose improved versions of the recently proposed Localization-aware ECE~\cite{saod} and show the efficacy of our method on these metrics as well.
Code is available at: \url{https://github.com/fiveai/detection_calibration}.

\end{abstract}

\section{Introduction}
\label{sec:intro}
Object detectors have been widely-used in a variety of safety-critical applications related to, but not limited to, autonomous driving \cite{geiger2012kitti, caesar2020nuscenes, sun2020scalability, grigorescu2019surveyAD, Cityscapes, bdd100k} and medical imaging \cite{yan2017deeplesion, kumar2020multiorgan, kumar2017pathology, jin2022anatomy}.
%
In addition to being accurate, their confidence estimates should also allow characterization of their error behaviour to make them reliable.
This feature, known as calibration, can enable a model to provide valuable information to subsequent systems playing crucial role in making safety-critical decisions \cite{karimi2020medical, mehrtash2020medical_calibration, lu2017associationlstm, sam, bewley2016SORT}.
Despite its importance, calibration of detectors is a relatively underexplored area in the literature and requires significant attention. 
Therefore, in this work, we focus on different aspects of the evaluation framework that is now being adopted by most recent works building calibrated detectors and discuss their pitfalls and propose fixes. Additionally, we tailor the well-known post-hoc calibration methods to improve the calibration of a given object detector (trained) with minimal effort.

\begin{figure*}[t]
        \captionsetup[subfigure]{}
        \centering
        \begin{subfigure}[b]{0.24\textwidth}
        \centering
            \includegraphics[width=\textwidth]{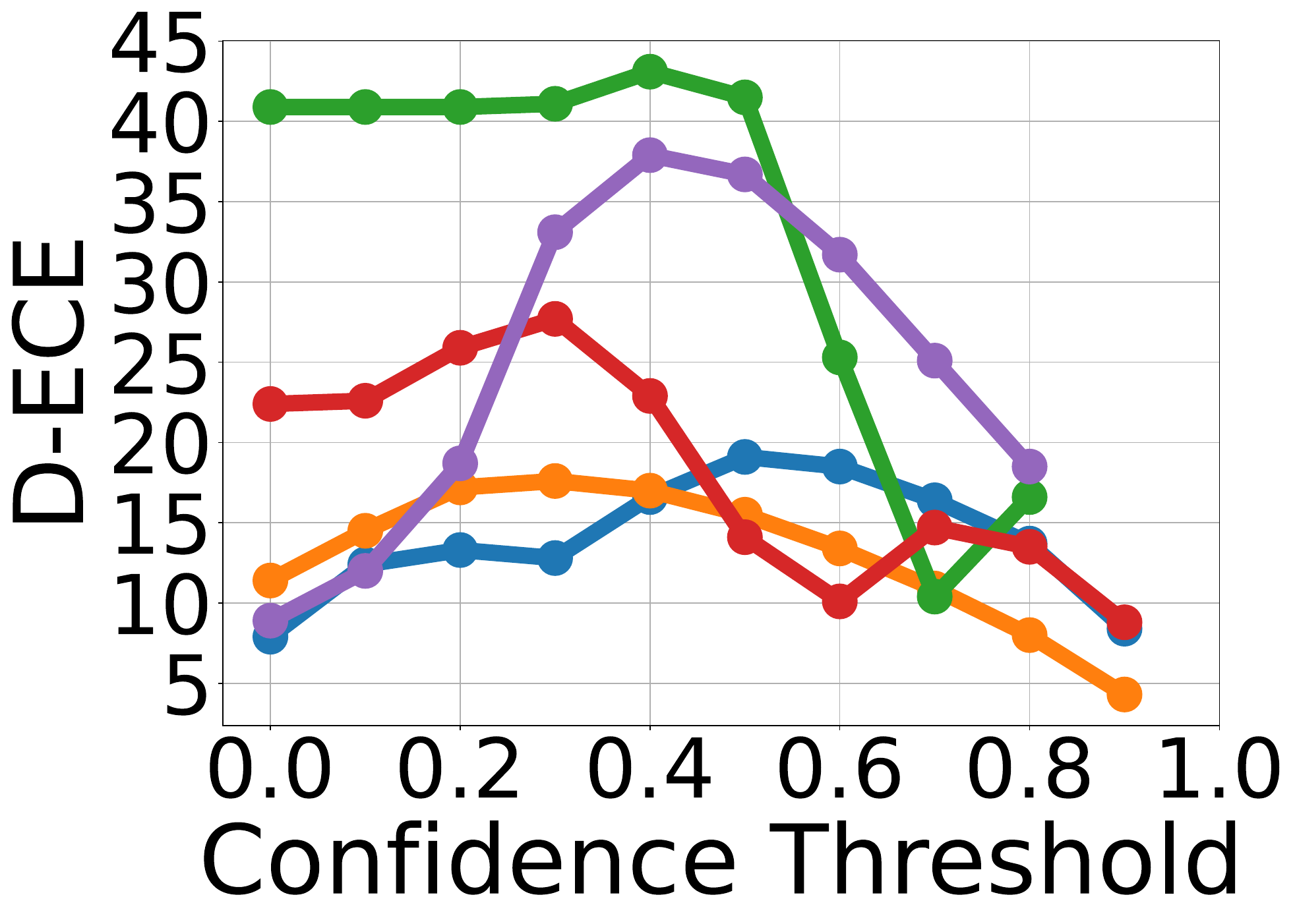}
            \caption{D-ECE}
        \end{subfigure}
        \begin{subfigure}[b]{0.24\textwidth}
        \centering
            \includegraphics[width=\textwidth]{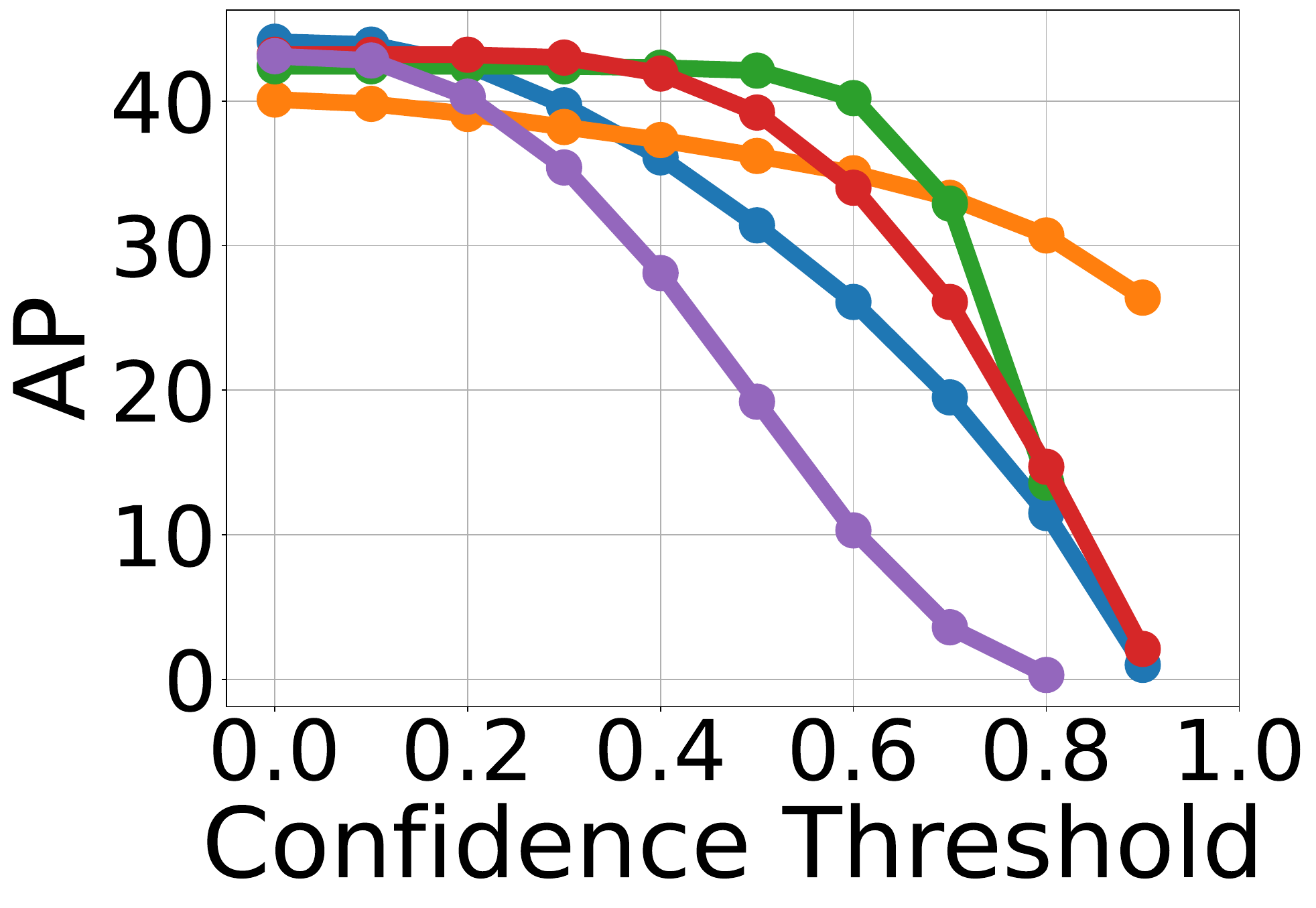}
            \caption{AP}
        \end{subfigure}
        \begin{subfigure}[b]{0.24\textwidth}
        \centering
            \includegraphics[width=\textwidth]{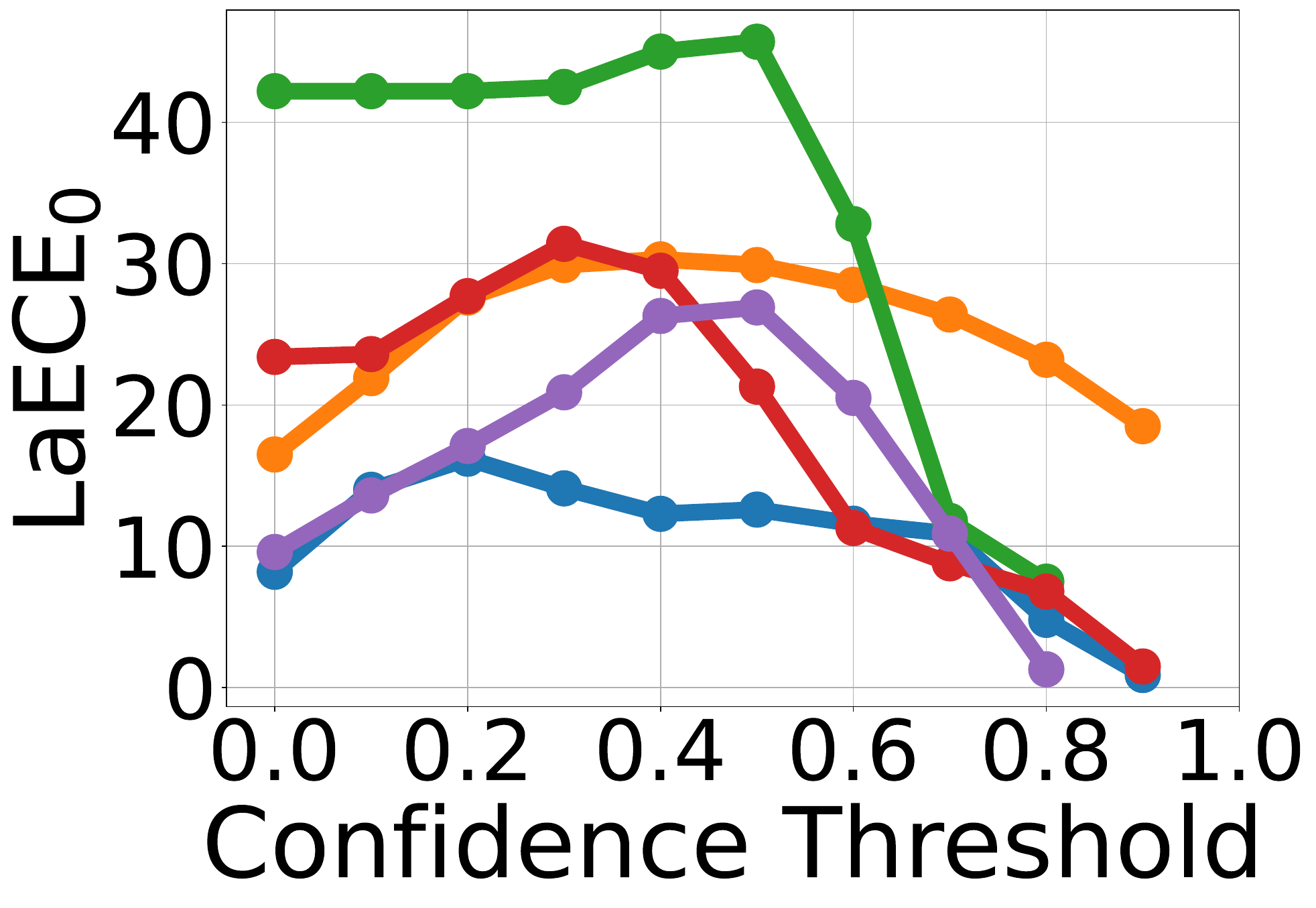}
            \caption{$\mathrm{LaECE_0}$}
        \end{subfigure}
        \begin{subfigure}[b]{0.24\textwidth}
        \centering
            \includegraphics[width=\textwidth]{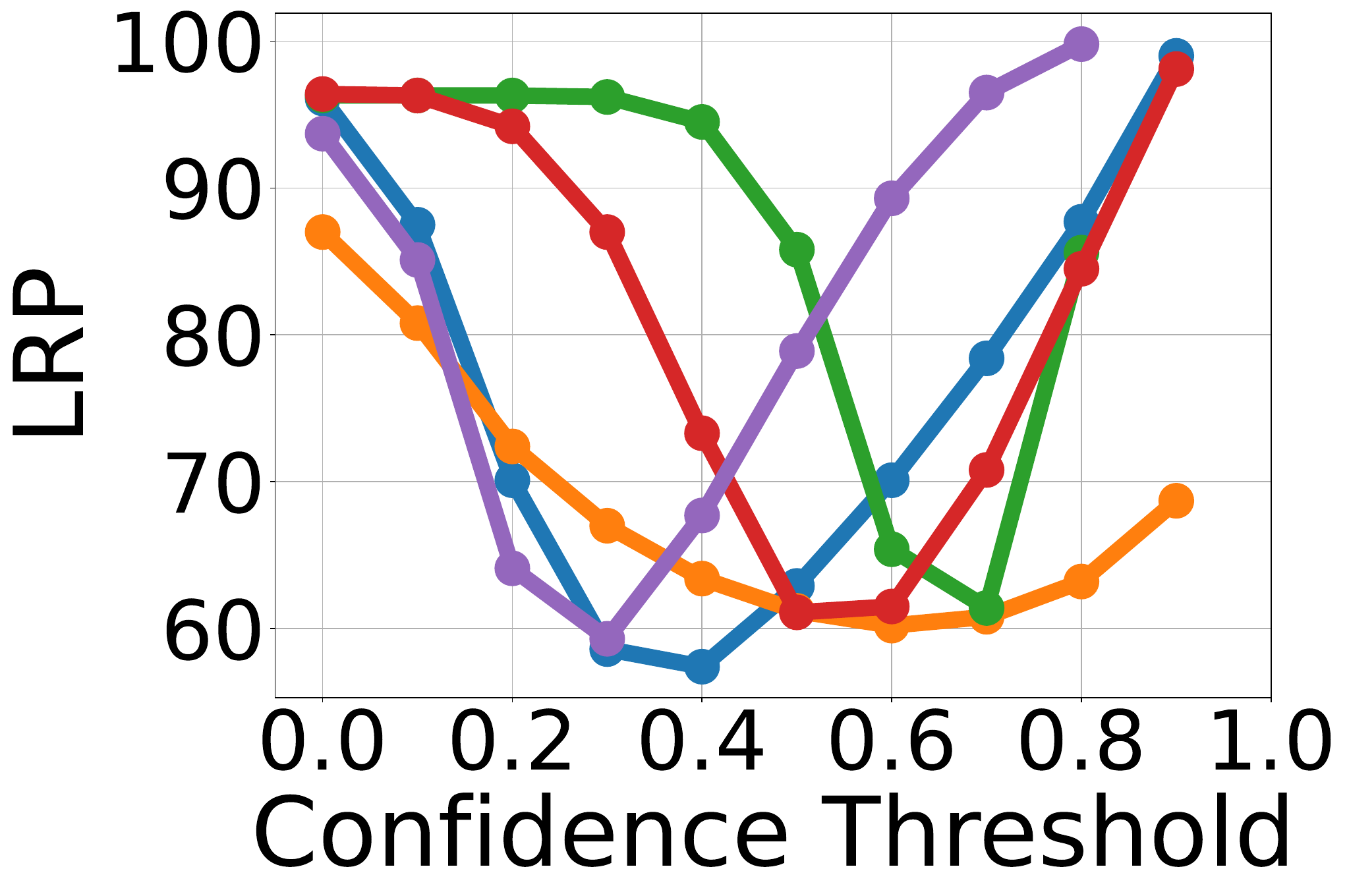}
            \caption{LRP}
        \end{subfigure}
        \caption{The performance of different detectors 
        over operating confidence thresholds on COCO \textit{minitest}. \textbf{\textcolor{orange}{Orange}}: Faster R-CNN, \textbf{\textcolor{ForestGreen}{Green}}: RS R-CNN, \textbf{\textcolor{Mulberry}{Purple}}: ATSS, \textbf{\textcolor{BrickRed}{Red}}: PAA, \textbf{\textcolor{BlueViolet}{Blue}}: D-DETR.
        All measures are lower better except AP. 
        It is not trivial to identify an operating threshold and compare detectors, especially when the common evaluation \cite{CalibrationOD,munir2022tcd, MCCL, munir2023bpc, munir2023caldetr}, combining D-ECE for calibration and AP for accuracy, is used. Instead, we use $\mathrm{LaECE_0}$ and \gls{LRP}. 
        }
        \label{fig:question}
\end{figure*}

Naturally, practitioners prefer detectors that perform well in terms of \textit{both accuracy and calibration}, which we refer to as joint performance.
However, unlike classification, choosing the best performing model is non-trivial for object detection.
This is because different detectors commonly yield detection sets with varying cardinalities for the same image, and this difference in population size is shown to affect the joint performance evaluation \cite{saod}.
Furthermore, when object detectors are used in practice, an operating threshold is normally chosen~\cite{karimi2020medical, mehrtash2020medical_calibration, lu2017associationlstm, sam, groundingdino, glip, bewley2016SORT}, and the choice of this threshold directly influences a detector's performance.
Thus, comparing the performance of a detector in terms of calibration or accuracy over different operating thresholds, as well as with different detectors, is not straightforward as illustrated in \cref{fig:question}.

We assert that a framework for joint evaluation should follow certain basic principles.
Firstly, the detectors should be evaluated on a thresholded set of detections to align with their practical usage. 
While doing so, the evaluation framework will require a principled \textit{model-dependent threshold selection} mechanism, as the confidence distribution of each detector can differ significantly \cite{LRPPAMI}. 
Secondly, the calibration evaluation should involve \textit{fine-grained information about the detection quality}. 
For example, if the confidence score represents the localisation quality of a detection, this provides more fine-grained information than only representing whether the object is detected or not.
Thirdly, \textit{the datasets should be properly-designed} for evaluation.
That is, the training, validation (val.) and \gls{ID} test splits should be sampled from the same underlying distribution, and additionally, the domain-shifted test splits --- which are crucial for safety-critical applications --- should be included.
Finally, \textit{baseline detectors and calibration methods must be trained properly}, as otherwise the evaluation might provide misleading conclusions.

There are three approaches in the literature attempting joint evaluation of accuracy and calibration as follows:
\begin{compactitem}
    \item[$\circ$] \textit{D-ECE-style} \cite{CalibrationOD,munir2022tcd, MCCL, munir2023bpc, munir2023caldetr}: thresholds the detections commonly from a confidence of $0.30$ to compute \gls{DECE} and use top-100 detections from each image for \gls{AP},
     \item[$\circ$] \textit{LaECE-style} \cite{saod}: enforces the detectors to be thresholded properly, and combine \gls{LaECE} with \gls{LRP} \cite{LRPPAMI},
     \item[$\circ$] \textit{CE-style} \cite{popordanoska2024CE}: thresholds the detections from a confidence score of $0.50$ to obtain \gls{CE} and \gls{AP}.
\end{compactitem}
As summarized in \cref{tab:limitations}, these evaluations do not adhere to the basic principles mentioned above.
To exemplify, \gls{DECE}-style evaluation --- the most common evaluation approach \cite{CalibrationOD,munir2022tcd, MCCL, munir2023bpc, munir2023caldetr} --- uses different operating thresholds for calibration and accuracy, which does not align well with the practical usage of detectors.
Also, using a fixed threshold for all detectors artificially promotes certain detectors.
To illustrate, while \gls{DECE}-style evaluation (threshold 0.3) ranks the green detector as the worst in \cref{fig:question}(a), the green one yields the best \gls{DECE} at $0.70$.
Besides, as shown in \cref{fig:question}(b), \gls{AP} is maximized at the confidence of $0$ (leading to too many detections with low confidences) for all the detectors, and thus \gls{AP} cannot be used to obtain a proper operating threshold \cite{LRPPAMI,saod}.
In terms of conveying fine-grained information, \gls{DECE} aims to align confidence with the precision only, which effectively ignores the localisation quality of the detections, a crucial performance aspect of object detection.
Finally, this type of evaluation also has limitations in terms of dataset splits and the chosen baselines as we explore in \cref{sec:pitfall}.

Having proper baseline calibration methods is also essential to monitor the progress in the field.
Recently proposed train-time calibration methods commonly employ an auxiliary loss term to regularize the confidence scores during training \cite{CalibrationOD,munir2022tcd, MCCL, munir2023bpc, munir2023caldetr}. 
Such methods are shown to be effective against the \gls{TS} \cite{calibration}, which is used as the only post-hoc calibration baseline.
Post-hoc calibrators are obtained on a held-out val. set, and hence can easily be applied to any off-the-shelf detector.
Despite their potential advantages, unlike for classification \cite{calibration, rahimi2022posthoc, ma2021metacal, hekler2023calibration, wang2021rethinkcalibration, zhang2023transferablecalibration, joy2023adaptive}, post-hoc calibration methods have not been explored for object detection sufficiently \cite{CalibrationOD,saod}.

\begin{table}[t]
    \centering
    \caption{Principles of joint performance evaluation of object detectors in terms of accuracy and calibration, and whether existing evaluation approaches violate them. 
    }
    \scalebox{0.88}{
    \begin{tabular}{c|c|c|c||c}
         \hline
         Principles of Joint Evaluation&\gls{DECE}-style
         \cite{CalibrationOD, MCCL, munir2023bpc, munir2022tcd, munir2023caldetr}
         &LaECE-style
         \cite{saod}
         &CE-style
         \cite{popordanoska2024CE}
         &\textit{Ours}\\ \hline \hline
         Model-dependent threshold selection& \xmark & \cmark & \xmark& \cmark  \\ \hline 
         Fine-grained confidence scores& \xmark & \xmark & \xmark & \cmark\\ \hline
         Properly-designed datasets&\xmark &\xmark & \xmark  & \cmark \\ \hline 
         Properly-trained detectors \& calibrators &\xmark &\cmark & \xmark  & \cmark \\ \hline 
    \end{tabular}
    }
    \label{tab:limitations}
\end{table}

In this paper, we introduce a joint evaluation framework which respects the aforementioned principles (\cref{tab:limitations}), and thus address the critical drawbacks of existing evaluation approaches.
That is, we first define $\mathrm{LaECE}_0$ and $\mathrm{LaACE}_0$, as novel calibration errors, each of which aims to align the detection confidence scores with their localisation qualities.
Specifically, the detectors respecting $\mathrm{LaECE}_0$ and $\mathrm{LaACE}_0$ provide quite informative confidence estimates about their behaviours.
We measure accuracy using \gls{LRP} \cite{LRPPAMI}, which requires a proper combination of \gls{FP}, \gls{FN} and localisation errors. 
Thereby requiring the detectors to be properly-thresholded as shown by the bell-like curves in \cref{fig:question}(d).
Also, we design three datasets with different characteristics, and introduce \gls{PS} as well as \gls{IR} as \textit{highly effective} post-hoc calibrators tailored to object detection. 
%
Our main contributions are:
\begin{compactitem}
    \item[$\circ$] We identify various quirks and assumptions in \gls{SOTA} methods in quantifying miscalibration of object detectors and show that they, if not treated properly, can provide misleading conclusions.
    \item[$\circ$] We introduce a framework for joint evaluation consisting of properly-designed datasets, evaluation measures tailored to practical usage of object detectors and baseline post-hoc calibration methods. We show that our framework addresses the drawbacks of existing approaches.
    \item[$\circ$] In contrast to the literature, we show that, if designed properly, post-hoc calibrators can significantly outperform the \gls{SOTA} training time calibration methods. To illustrate, on the common COCO benchmark, D-DETR with our \gls{IR} calibrator outperforms the \gls{SOTA} Cal-DETR \cite{munir2023caldetr} significantly: (i) by more than $7$ points in terms of \gls{DECE} and (ii) $\sim 4$ points in terms of our challenging $\mathrm{LaECE}_0$ 
\end{compactitem}
\section{Background and Notation} \label{sec:relatedwork}

\textbf{Object Detectors and Evaluating their Accuracy}
Denoting the set of $M$ objects in an image $X$ by $\{b_i, c_i\}^M$ where $b_i \in \mathbb{R}^{4}$ is a bounding box and $c_i \in  \{1,\dots, K\}$ is its class; an object detector produces the bounding box $\hat{b}_i$, the class label $\hat{c}_i$ and the confidence score $\hat{p}_i$  for the objects in $X$, i.e., $f(X) = \{\hat{c}_i, \hat{b}_i, \hat{p}_i\}^N$ with $N$ being the number of predictions.
During evaluation, each detection is first labelled as a \gls{TP} or a  \gls{FP} using a matching function $\psi(\cdot)$ relying on an \gls{IoU} threshold $\tau$ to validate \glspl{TP}.
%
We assume $\psi(i)$ returns the index of the object that a \gls{TP} $i$ matches to; else $i$ is a \gls{FP} and $\psi(i) = -1$.
Then, \gls{AP}~\cite{COCO,LVIS,PASCAL}, the common accuracy measure, corresponds to the area under the \gls{PR} curve.
Though widely-used, \gls{AP} has been criticized recently from different aspects~\cite{LRPPAMI,TIDE,Trustworthy,LRP,OptCorrCost}.
To illustrate, \gls{AP} is maximized when the number of detections increases \cite{saod} as shown in \cref{fig:question}(b). 
Therefore, \gls{AP} does not help choosing an operating threshold, which is critical for practical deployment.
As an alternative, \gls{LRP}~\cite{LRP,LRPPAMI}
combines the numbers of \gls{TP}, \gls{FP}, \gls{FN} with the localisation error of the detections, which are denoted by $\mathrm{N_{TP}}$, $\mathrm{N_{FP}}$, $\mathrm{N_{FN}}$ and $\mathcal{E}_{loc}(i) \in [0,1]$ respectively:
{
\begin{align}\label{eq:LRPdefcompact}
     \mathrm{LRP} = \frac{1}{\mathrm{N_{FP}} +\mathrm{N_{FN}}+\mathrm{N_{TP}}}\left(\mathrm{N_{FP}} +\mathrm{N_{FN}} + \sum \limits_{\psi(i) > 0} \mathcal{E}_{loc}(i) \right).
\end{align}
}
Unlike \gls{AP}, \gls{LRP} requires the detection set to be thresholded properly as both \glspl{FP} and \glspl{FN} are penalized in Eq. \eqref{eq:LRPdefcompact}.

\textbf{Evaluating the Calibration of Object Detectors}
The alignment of accuracy and confidence of a model, termed calibration, is extensively studied for classification ~\cite{calibration,AdaptiveECE,verifiedunccalibration,rethinkcalibration,FocalLoss_Calibration,calibratepairwise}.
That is, a classifier is \textit{calibrated} if its accuracy is $p$ for the predictions with confidence of $p$ for all $p \in [0,1]$. 
For object detection, \cite{CalibrationOD} extends this definition to enforce that the confidence matches the precision of the detector,
$\mathbb{P}(\hat{c}_i = c_i | \hat{p}_i)  = \hat{p}_i, \forall \hat{p}_i \in [0,1],$
where $\mathbb{P}(\hat{c}_i = c_i | \hat{p}_i)$ is the precision.
Then, discretizing the confidence space into $J$ bins, \gls{DECE} is
\begin{align}
\label{eq:dece_}
   \text{\gls{DECE}} 
    = \sum_{j=1}^{J} \frac{|\hat{\mathcal{D}}_j|}{|\hat{\mathcal{D}}|} \left\lvert \bar{p}_{j} - \mathrm{precision}(j)  \right\rvert,
\end{align}
where $\hat{\mathcal{D}}$ and $\hat{\mathcal{D}}_j$ are the set of all detections and the detections in the $j$-th bin, and $\bar{p}_{j}$ and $\mathrm{precision}(j)$ are the average confidence and the precision of the detections in the $j$-th bin.
Alternatively, considering that object detection is a joint task of classification and localisation, \gls{LaECE} \cite{saod} aims to match the confidence with the product of precision and average \gls{IoU} of \glspl{TP}. 
Also, to prevent certain classes from dominating the error, \gls{LaECE} is introduced as a class-wise measure.
Using superscript $c$ to refer to each class and $\bar{\mathrm{IoU}}^{c}(j)$ as the average \gls{IoU} of $\hat{\mathcal{D}}^{c}_j$, \gls{LaECE} is defined as:
\begin{align}
\label{eq:laece__}
   \mathrm{LaECE} 
    = \frac{1}{K} \sum_{c=1}^{K} \sum_{j=1}^{J} \frac{|\hat{\mathcal{D}}^{c}_j|}{|\hat{\mathcal{D}}^{c}|} \left\lvert \bar{p}^{c}_{j} - \mathrm{precision}^{c}(j) \times \bar{\mathrm{IoU}}^{c}(j)  \right\rvert.
\end{align}

\textbf{Calibration Methods in Object Detection}
The existing methods for calibrating object detectors can be split into two groups:

\textit{(1) Training-time calibration approaches} \cite{munir2022tcd, MCCL, munir2023bpc, munir2023caldetr, popordanoska2024CE} regularize the model to yield calibrated confidence scores during training, which is generally achieved by an additive auxiliary loss.

\textit{(2) Post-hoc calibration methods}
use a held-out val. set to fit a calibration function that maps the predicted confidence to the calibrated confidence.
Specifically, \gls{TS} \cite{calibration} is the only method considered as a baseline for recent training time methods \cite{munir2022tcd, MCCL, munir2023bpc, popordanoska2024CE}.
As an alternative, \gls{IR} \cite{zadrozny2002transforming} is used within a limited scope for a specific task called Self-aware Object Detection \cite{saod}.
Furthermore, its effectiveness neither on a wide range of detectors nor against existing training-time calibration approaches has yet been investigated.

\section{Analysis of the Common \gls{DECE}-style Evaluation}
\label{sec:pitfall}

\begin{wrapfigure}{r}{6cm}
        \centering
        \includegraphics[width=0.35\textwidth]{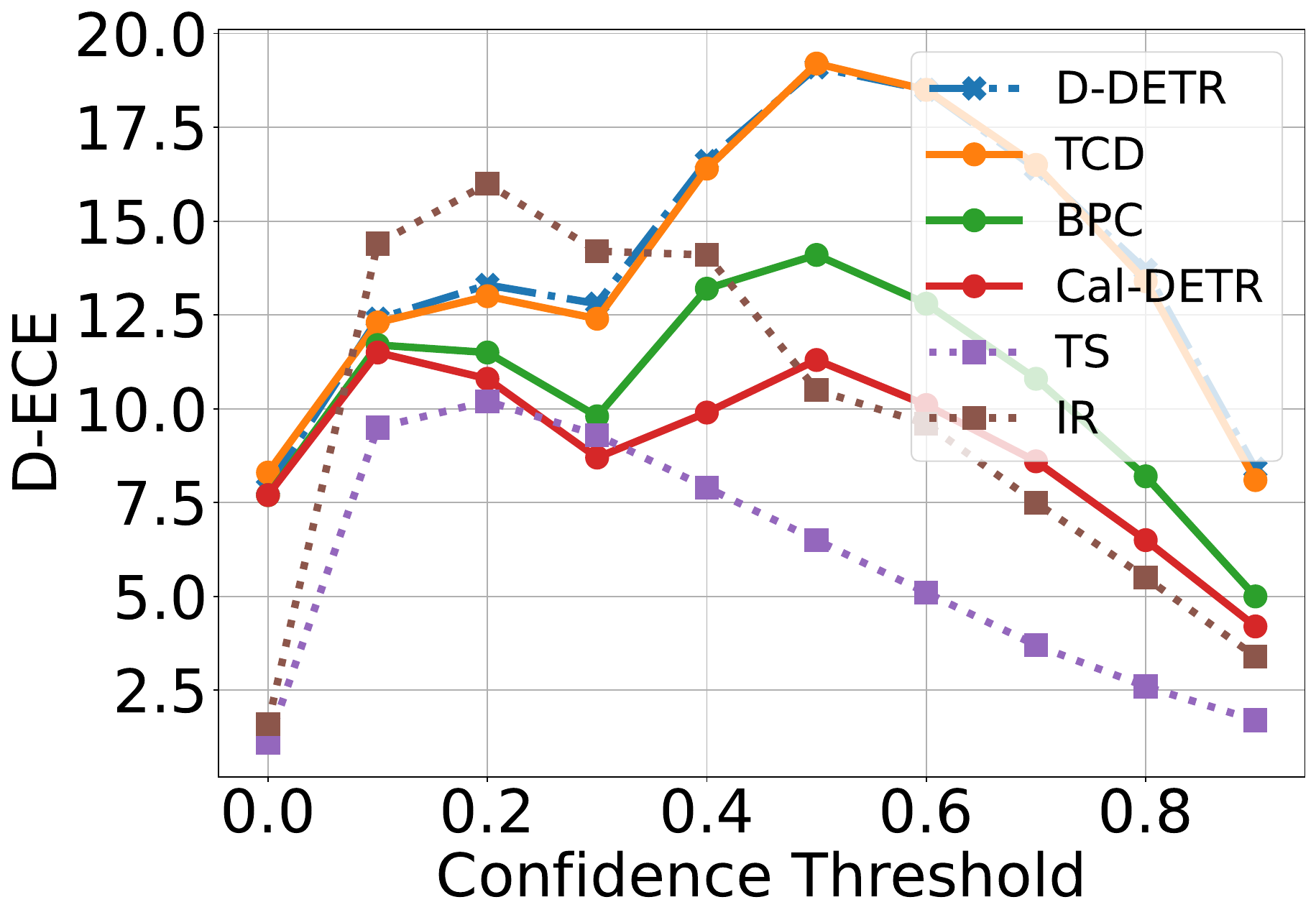}
        \caption{Comparison of calibration methods in terms of \gls{DECE} on COCO \textit{mini-test} using D-DETR \cite{DDETR}. Post-hoc calibrators \gls{TS} and \gls{IR} are obtained on a subset of Objects365 \cite{Objects365} following \gls{DECE}-style evaluation. 
        }
        \label{fig:curves}
        \vspace{-2.5ex}
\end{wrapfigure}
%
\gls{DECE}-style evaluation is the most common evaluation approach adopted by several methods \cite{CalibrationOD,munir2022tcd, MCCL, munir2023bpc, munir2023caldetr}. For that reason, here we provide a comprehensive analysis of this evaluation approach and analyse the \gls{LaECE}-style and \gls{CE}-style evaluations in App. \ref{app:analyses}. 
Our analyses, based on the principles outlined in \cref{sec:intro} and \cref{tab:limitations}, show that all approaches have notable drawbacks.

\textbf{1. Model-dependent threshold selection.} As \gls{AP} is obtained using the top-100 detections and \gls{DECE} is computed on detections thresholded above $0.30$, \gls{DECE}-style evaluation uses two different detection sets.
This inconsistency is not reflective of how detectors are used in practice.
Also, we observe that a fixed threshold of $0.30$ for evaluating the calibration induces a bias for certain detectors. 
To illustrate, we compare the performance of different calibration methods over different thresholds in \cref{fig:curves}, where Cal-DETR \cite{munir2023caldetr} performs the best only for the threshold $0.30$ and the post-hoc \gls{TS} significantly outperforms it on all other thresholds. 
Therefore, this method of evaluation is sensitive to the choice of threshold, leading to ambiguity on the best performing method.

\begin{figure*}[t]
        \captionsetup[subfigure]{}
        \centering
        \begin{subfigure}[b]{0.325\textwidth}
            \includegraphics[width=\textwidth]{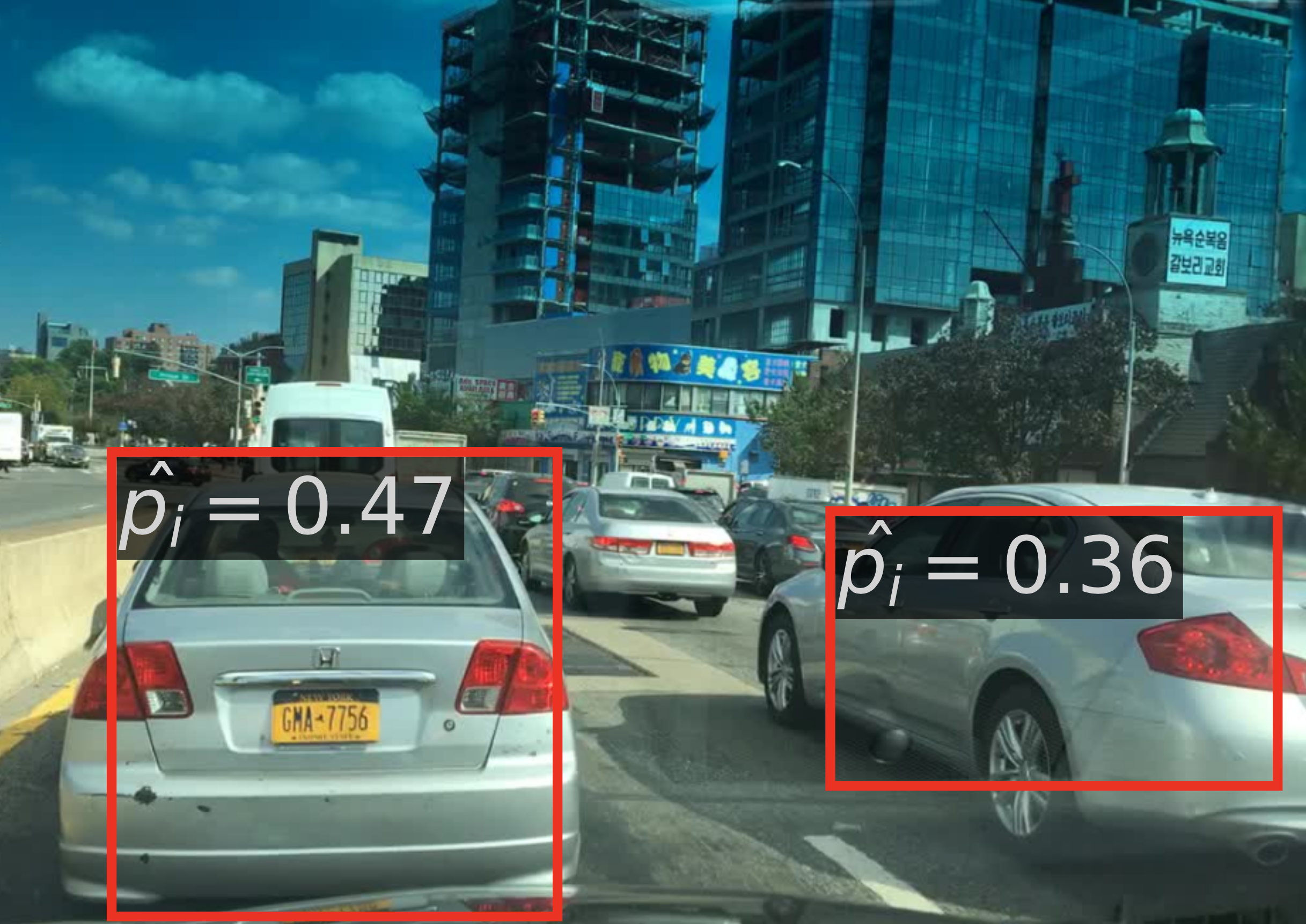}
            \caption{Uncalibrated detections}
        \end{subfigure}
        \begin{subfigure}[b]{0.325\textwidth}
            \includegraphics[width=\textwidth]{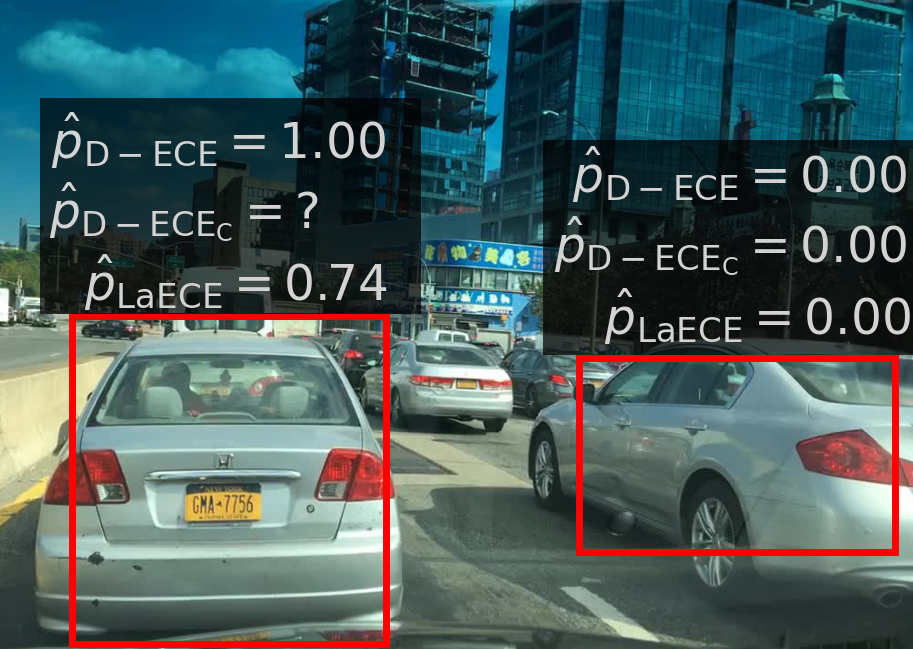}
            \caption{Calibrated detections}
        \end{subfigure}
        \begin{subfigure}[b]{0.325\textwidth}
            \includegraphics[width=\textwidth]{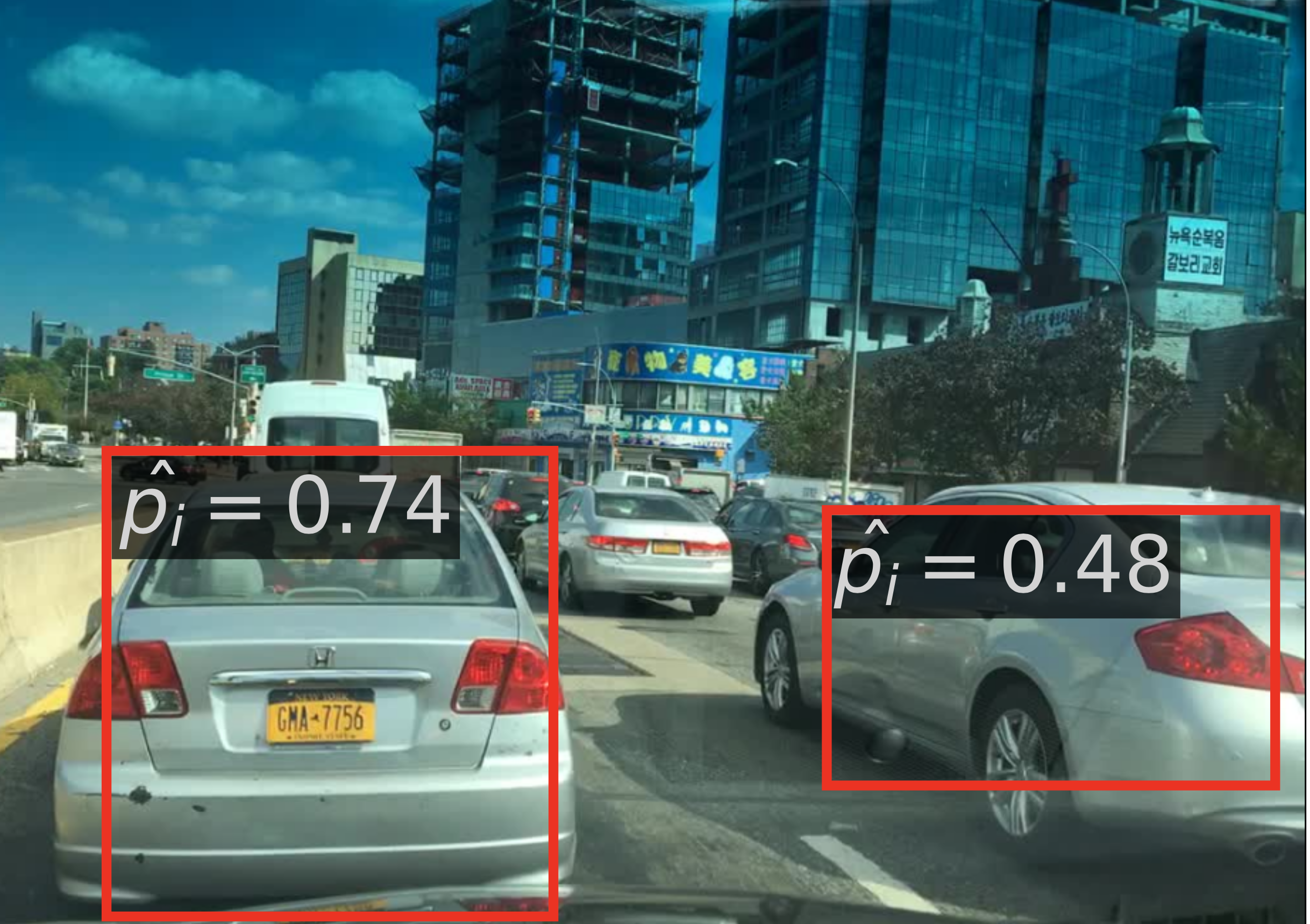}
            \caption{Calibrated detections (Ours)}
        \end{subfigure}
        \begin{subfigure}[b]{0.32\textwidth}
            \includegraphics[width=\textwidth]{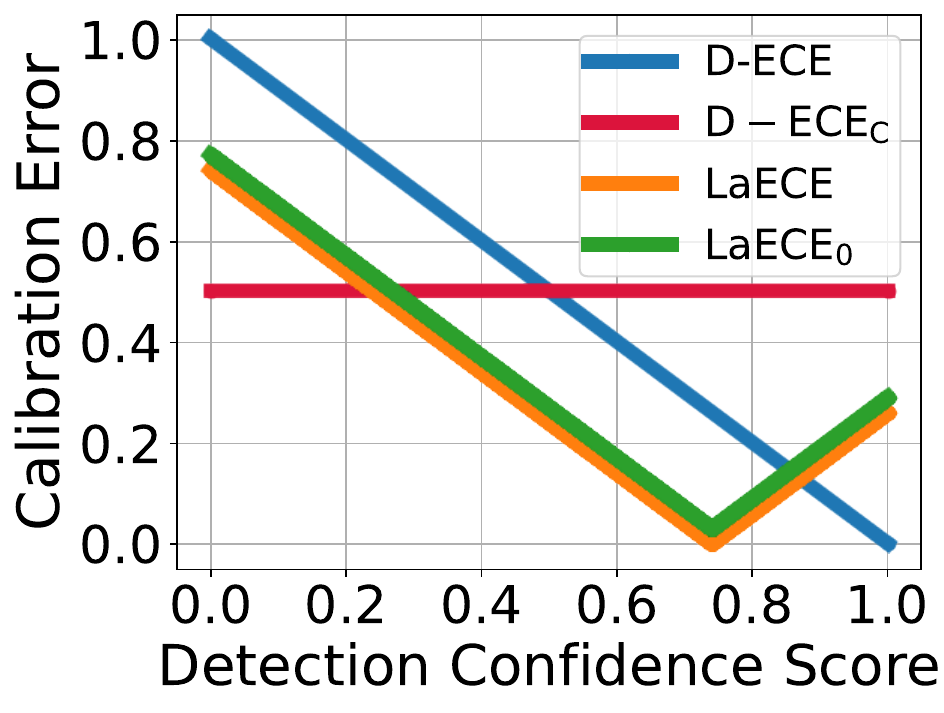}
            \caption{\gls{TP} for $\tau=0.5$}
        \end{subfigure}
        \begin{subfigure}[b]{0.32\textwidth}
            \includegraphics[width=\textwidth]{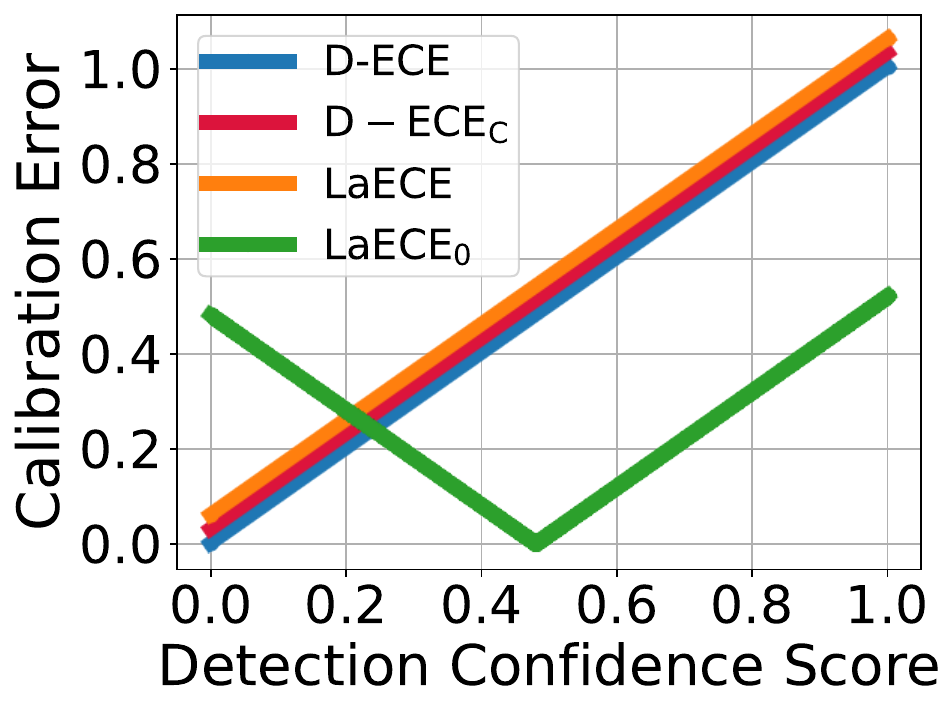}
            \caption{\gls{FP} for $\tau=0.5$, \gls{TP} for $\tau=0$}
        \end{subfigure}
        \begin{subfigure}[b]{0.32\textwidth}
            \includegraphics[width=\textwidth]{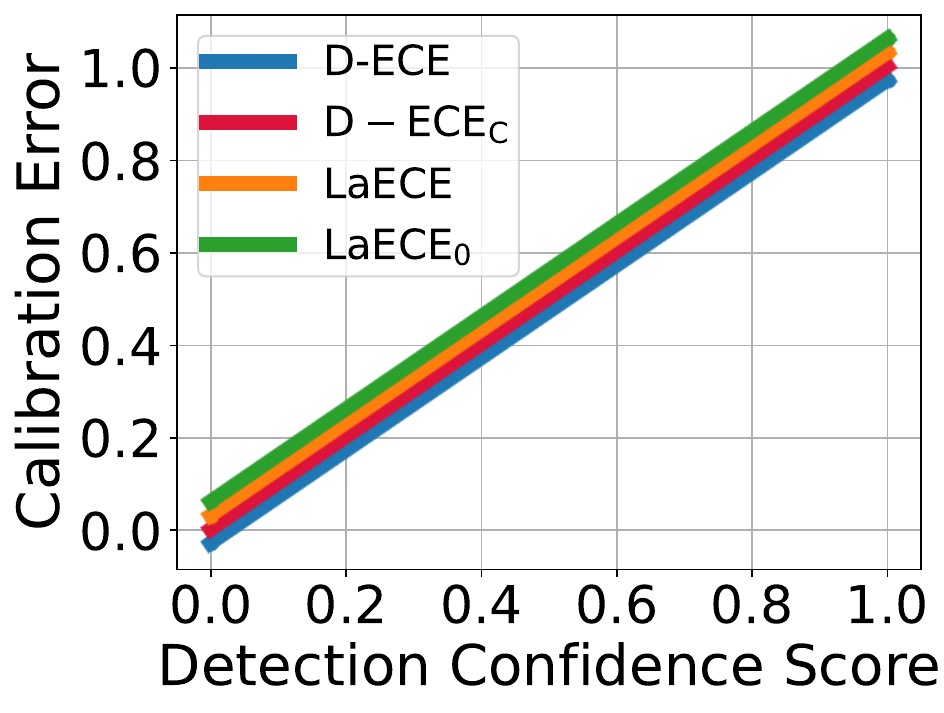}
            \caption{\gls{FP} for $\tau=0$}
        \end{subfigure}
        \caption{A pictorial comparison of the different calibration errors. (a) Uncalibrated detections 
        of D-DETR on an image from \cite{bdd100k}. The detections on the left and right have \glspl{IoU} of $0.74$ and $0.48$ with the objects. 
        (b) Calibrated detections in terms of \gls{DECE} and $\mathrm{LaECE}$ using $\tau=0.50$, and $\mathrm{\text{\gls{DECE}}_C}$, COCO-style \gls{DECE} as in \cite{popordanoska2024CE}. $\mathrm{D-ECE_C}=?$ as calibration error does not have a global minimum as shown in (d). (c) Calibrated detections in terms of $\mathrm{LaECE_{0}}$ and $\mathrm{LaACE_{0}}$ in which confidence matches \gls{IoU}. (d-f) Calibration errors for different types of detections, for which $\mathrm{LaACE_{0}}$ behave the same as $\mathrm{LaECE_{0}}$, hence excluded for clarity. App. \ref{app:analyses} presents the details.}
        \label{fig:images}
\end{figure*}

\textbf{2. Fine-grained confidence scores.} Manipulating Eq. \eqref{eq:dece_}, we show in App. \ref{app:analyses} that \gls{DECE} for the $j$-th bin can be expressed as,
\begin{align}
\label{eq:DECEbin}
  &\Bigg\lvert \sum_{\substack{\hat{b}_i \in \hat{\mathcal{D}}_j, \psi(i) > 0}} \big( \hat{p}_i - 1 \big) + \sum_{\substack{\hat{b}_i \in \hat{\mathcal{D}}_j, \psi(i) = -1}} \hat{p}_i  \Bigg\rvert.
\end{align}
Eq. \eqref{eq:DECEbin} implies that \gls{DECE} is minimized when the confidence scores $\hat{p}_i$ of \glspl{TP} are $1$ and those of \glspl{FP} are $0$, which is also how the prediction-target pairs are usually constructed to train post-hoc \gls{TS} \cite{CalibrationOD,munir2022tcd, MCCL, munir2023bpc, munir2023caldetr}. 
Even if the detector is perfectly calibrated for these binary targets, the confidence scores do not provide information about localisation quality as illustrated by binary-valued $\hat{p}_{\mathrm{\text{\gls{DECE}}}}$ for both detections in \cref{fig:images}(b). 
Also, Popordanoska et al. \cite{popordanoska2024CE} utilise \gls{DECE} in a COCO-style manner, that is they average \gls{DECE} over different \gls{TP} validation \gls{IoU} thresholds similar to COCO-style \gls{AP} \cite{COCO}.
However, we observe that this way of using \gls{DECE} can promote ambiguous confidence scores.
As an example, given two \gls{IoU} thresholds $\tau_1$ and $\tau_2$, a detection $\hat{b}_i$ with $\tau_1 \leq \mathrm{IoU}(\hat{b}_i, b_{\psi(i)}) < \tau_2$ is a \gls{TP} for $\tau_1$ but a \gls{FP} for $\tau_2$.
Thus, given Eq. \eqref{eq:DECEbin}, it follows that $\hat{b}_i$ has contradictory confidence targets for $\tau_1$ and $\tau_2$.
This is illustrated in \cref{fig:images}(d) in which $\mathrm{\text{\gls{DECE}}_C}$ (red line) remains constant regardless of the confidence.
Thus, using \gls{DECE} (or another calibration measure) in this way should be avoided.

\begin{table}[t]
\parbox{.48\linewidth}{
\centering
    \caption{Effect of using domain-shifted val. set on \gls{IR} calibrator. Results are reported on COCO-\textit{minitest}. Val. set is N/A for uncalibrated D-DETR and training time calibration method Cal-DETR.}
    \vspace{-1.5ex}
    \scalebox{0.95}{\begin{tabular}{c|c|c|c}
         \hline
         Method&Val set&\gls{DECE}&AP $\uparrow$ \\ \hline
         D-DETR&N/A&$12.8$& $44.1$\\ \hline
         Cal-DETR&N/A&$8.7$& $44.4$\\ \hline
         IR&Objects365&$14.2$& $44.1$\\ \hline
         IR&COCO&$\mathbf{1.3}$ \imp{7.4}& $44.1$\\ \hline 
    \end{tabular}}
    \label{tab:domainshift}}
\hfill
\parbox{.48\linewidth}{
\centering
\caption{COCO training settings are commonly adopted while training D-DETR on Cityscapes. When trained with larger images and longer, DETR performs slightly better than Cal-DETR.}
\vspace{-1.5ex}
    \scalebox{0.95}{\begin{tabular}{c|c|c|c}
         \hline
         Method&Training Style&\gls{DECE}&AP$\uparrow$\\ \hline
         D-DETR\cite{munir2023caldetr}&COCO&$13.8$& $26.8$ \\ \hline
         Cal-DETR\cite{munir2023caldetr}&COCO&$8.4$&$28.4$\\ \hline
         Cal-DETR&Cityscapes&$4.0$& $34.9$\\ \hline
         D-DETR&Cityscapes&$\mathbf{2.9}$&$\mathbf{36.1}$\\ \hline
    \end{tabular}}
    \label{tab:proper_training_cityscapes}
}
\end{table}

\blockcomment{\gls{DECE} does not provide fine grained information as it dictates a confidence score to be either 1 or 0. We show in App. that \gls{DECE} for bin $j$ reduces to
{\small
\begin{align}
\label{eq:DECE_bin}
  \Bigg\lvert \sum_{\substack{\hat{b}_i \in \hat{\mathcal{D}}_j \\ \psi(i) > 0}} \big( \hat{p}_i - 1) + \sum_{\substack{\hat{b}_i \in \hat{\mathcal{D}}_j \\ \psi(i) \leq 0}} \hat{p}_i  \Bigg\rvert.
\end{align}
}
}

\textbf{3. Properly-designed datasets.} In the literature, the val. set to obtain the post-hoc calibrator is typically taken from a different dataset than the \gls{ID} dataset \cite{munir2022tcd, MCCL, munir2023bpc, munir2023caldetr}. Specifically, the post-hoc calibrators are obtained on a subset from from Objects365 \cite{Objects365} and BDD100K \cite{bdd100k} for the models trained with COCO \cite{COCO} and Cityscapes \cite{Cityscapes} respectively. Hence, as expected, a different dataset inevitably induces domain shift, affecting the performance of the post-hoc calibrator \cite{domainshiftposthoc}. 
To show that, following existing approaches, we obtain an \gls{IR} calibrator \cite{saod} on Objects365 and compare it with the one obtained on the \gls{ID} val. set in terms of \gls{DECE}-style evaluation. 
\cref{tab:domainshift} shows that the latter \gls{IR} now outperforms (i) the former one by $\sim 11$ \gls{DECE} and (ii) \gls{SOTA} Cal-DETR \cite{munir2023caldetr} by $7.4$ \gls{DECE}, showing the importance of dataset design for proper evaluation.

\textbf{4. Properly-trained detectors and calibrators.}  
Though Cityscapes is commonly used in the literature \cite{munir2022tcd, MCCL, munir2023bpc, munir2023caldetr}, the models trained on this dataset follow COCO-style training.
Specifically, D-DETR \cite{DDETR} was trained on Cityscapes for only 50 epochs though the training set of Cityscapes is $\sim 40 \times$ smaller than that of COCO ($3$K vs. $\sim 118$K).
We now tailor the training of D-DETR for Cityscapes by (i) $4 \times$ longer training considering the smaller training set and (ii) increasing the training image scale considering the original resolution following \cite{Cityscapes,mmdetection}.
We kept all other hyperparameters as they are for both Cal-DETR and D-DETR, and App. \ref{app:analyses} presents the details.
\cref{tab:proper_training_cityscapes} shows that, once trained in this setting, D-DETR performs better than Cal-DETR in terms of both accuracy and calibration. 
In the next section we discuss that baseline post-hoc calibration methods are not tailored for object detection either.

\section{A Framework for Joint Evaluation of Object Detectors}
We now present our evaluation approach that respects to the principles in \cref{sec:intro}.

\subsection{Towards Fine-grained Calibrated Detection Confidence Scores}
Calibration refers to the alignment of accuracy and confidence of a model.
Therefore, for an object detector to be calibrated, its confidence should respect both classification and localisation accuracy.
We discussed in \cref{sec:pitfall} that \gls{DECE}, as the common calibration measure, only considers the precision of a detector, thereby ignoring its localisation performance (Eq. \eqref{eq:dece_}).
\gls{LaECE} \cite{saod}, defined in Eq. \eqref{eq:laece__} as an alternative to \gls{DECE}, enforces the confidence scores to represent the product of precision and average \gls{IoU} of \glspl{TP}.
Thus, \gls{LaECE} considers \glspl{IoU} of only \glspl{TP}, and effectively ignores the localisation qualities of detections if their \gls{IoU} is less than the \gls{TP} validation threshold $\tau>0$.
We assert that this selection mechanism based on \gls{IoU} unnecessarily limits the information conveyed by the confidence score.
We illustrate this on the right car in \cref{fig:images}(b) for which \gls{LaECE} requires a target confidence of $0$ ($\hat{p}_{\mathrm{\text{\gls{LaECE}}}}=0$) as its \gls{IoU} is less than $\tau=0.50$.
However, instead of conveying a $0$ confidence and implying no detection, representing its \gls{IoU} by $\hat{p}_{i}$ provides additional information.
Hence, we propose using $\tau=0$, in which case the calibration criterion of \gls{LaECE} reduces to,
\begin{align}\label{eq:elce_criterion}
    \mathbb{E}_{\hat{b}_i \in B_i(\hat{p}_i)}[\mathrm{IoU}(\hat{b}_i, b_{\psi(i)})] = \hat{p}_i, \forall \hat{p}_i \in [0,1],
\end{align}
where we define $\mathrm{IoU}(\hat{b}_i, b_{\psi(i)})=0$ for \glspl{FP} when $\tau=0$, $B_i(\hat{p}_i)$ is the set of boxes with the confidence of $\hat{p}_i$ and $b_{\psi(i)}$ is the ground-truth box that $\hat{b}_i$ matches with.
To derive the calibration error for Eq. \eqref{eq:elce_criterion}, we follow \gls{LaECE} by using $J=25$ equally-spaced bins and averaging over class-wise errors and define,
\begin{align}
\label{eq:elce}
   \mathrm{LaECE}_0 
   & = \frac{1}{K} \sum_{c = 1}^K \sum_{j=1}^{J} \frac{|\hat{\mathcal{D}}^{c}_j|}{|\hat{\mathcal{D}}^{c}|} \left\lvert \bar{p}^{c}_{j} - \bar{\mathrm{IoU}}^{c}(j)  \right\rvert,
\end{align}
where $\hat{\mathcal{D}}^{c}$ and $\hat{\mathcal{D}}_j^{c}$ denote the set of all detections and those in $j$th bin respectively, $\bar{p}_{j}^{c}$ is the average confidence score and $\bar{\mathrm{IoU}}^{c}(j)$ is the average \gls{IoU} of detections in the $j$-th bin for class $c$, and the subscript $0$ refers to the chosen $\tau$ which is $0$.
Furthermore, similar to the classification literature \cite{FocalLoss_Calibration,AdaptiveECE}, we define \gls{LaACE} using an adaptive binning approach in which the number of detections in each bin is equal.
In order to capture the model behaviour precisely, we adopt the extreme case in which each bin has only one detection, resulting in an easy-to-interpret measure which corresponds to the mean absolute error between the confidence and the \gls{IoU},
\begin{align}
\label{eq:laace}
  \mathrm{LaACE}_0 
  & = \frac{1}{K} \sum_{c = 1}^K \sum_{i =1}^{|\hat{\mathcal{D}}^{c}|} \frac{1}{|\hat{\mathcal{D}}^{c}|} \left\lvert \hat{p}_{i} - \mathrm{IoU}(\hat{b}_i, b_{\psi(i)})  \right\rvert.
\end{align}
As we show in App. \ref{app:method},  $\mathrm{LaECE}_0$ and $\mathrm{LaACE}_0$ are both minimized when $\hat{p}_{i}=\mathrm{IoU}(\hat{b}_i, b_{\psi(i)})$ for all detections, which is also a necessary condition for $\mathrm{LaACE}_0$.
Hence, as illustrated on the right car in \cref{fig:images}(c) and (e), $\mathrm{LaECE}_0$ and $\mathrm{LaACE}_0$ requires conveying more fine-grained information compared to other measures.

\subsection{Model-dependent Thresholding for Proper Joint Evaluation}
In practice, object detectors employ an operating threshold to preferably output only \glspl{TP} with high recall.
However, \gls{AP} as the common performance measure does not enable cross-validating such a threshold as it is maximized when the recall is maximized despite a drop in precision \cite{LRPPAMI,saod}.
This can be observed in \cref{fig:question}(b) where \gls{AP} consistently decreases as the confidence threshold increases.
Alternatively, \gls{LRP} (Eq. \ref{eq:LRPdefcompact}) prefers detectors with high precision, recall and low localisation error as illustrated by the bell-like curves in \cref{fig:question}(d).
This is because, unlike \gls{AP}, \gls{LRP} severely penalises detectors with low recall or precision, making it a perfect fit for our framework.
As a result, we combine $\mathrm{LaECE}_0$ and $\mathrm{LaACE}_0$ with \gls{LRP} and require each model to be thresholded properly.

\begin{table}[t]
    \centering
    \small
    \caption{Datasets for evaluating object detection and instance segmentation methods.}
    \scalebox{1.0}{\begin{tabular}{c|c|c|c|c}
         \hline
         Type&Train set&Val set& \gls{ID} test set&Domain-shifted test set\\ \hline \hline
          Common Objects&COCO train & COCO minival & COCO minitest&COCO minitest-C, Obj45K  \\  
          Autonomous Driving&CS train & CS minival & CS minitest & CS minitest-C, Foggy-CS\\  
         Long-tailed Objects&LVIS train&LVIS minival&LVIS minitest&LVIS minitest-C\\ \hline
    \end{tabular}}
    \label{tab:datasets}
\end{table}

\subsection{Properly-designed Datasets}
We curate three datasets summarized in \cref{tab:datasets}: (i) COCO \cite{COCO} including common daily objects; (ii) Cityscapes \cite{Cityscapes} with autonomous driving scenes; and (iii) LVIS \cite{LVIS}, a challenging dataset focusing on the calibration of long-tailed detection.
For each dataset, we ensure that train, val. and \gls{ID} test sets are sampled from the same distribution, and include domain-shifted test sets.
As these datasets do not have public labels for test sets, we randomly split their val. sets into two as minival and minitest similar to \cite{saod,MOCAE,RegressionUncOD}.
In such a way, we provide \gls{ID} val. sets to enable obtaining post-hoc calibrators and the operating thresholds properly.
For domain-shifted test sets, we apply common corruptions \cite{hendrycks2019robustness} to the \gls{ID} test sets, and 
include Obj45K \cite{saod,Objects365} and Foggy Cityscapes \cite{sakaridis2018foggycs} as more realistic shifts.
Our datasets also have mask annotations and hence they can be used to evaluate instance segmentation methods. 
App. \ref{app:method} includes further details.

\subsection{Baseline Post-hoc Calibrators Tailored to Object Detection}
\label{sec:posthoc}
It is essential to develop post-hoc calibration methods tailored to object detection, which has certain differences from the classification task.
However,  existing methods \cite{munir2022tcd, MCCL, munir2023bpc, munir2023caldetr, popordanoska2024CE} use only \gls{TS} as a baseline without considering the peculiarities of detection.
Specifically, a single temperature parameter $T$ is learned to adjust the predictive distribution while the detection confidence score $\hat{p}_i$ is commonly assumed to be a Bernoulli random variable \cite{CalibrationOD}.
On the other hand, \gls{PS}, which fits both a scale and a shift parameter, is the widely-accepted calibration approach when the underlying distribution is Bernoulli \cite{plat2000probabilistic, calibration}.
Also, how to construct a useful subset of the detections to train the post-hoc calibrators has not been explored.
To address these shortcomings, we present (i) Platt Scaling in which the bias term makes a notable difference in the performance, and (ii) Isotonic Regression by modeling the calibration as a regression problem.
Before introducing them, we now present an overview on how we determine the set of detections to train the calibrators. 

\textbf{Overview} We obtain post-hoc calibrators on a held out val. set using the detections that are similar to those seen at inference to prevent low-scoring detections from dominating the training of the calibrator.
To do so, we cross-validate a calibration threshold $\bar{u}^c$ for each class $c$ and train a class-specific calibrator $\zeta^c:[0,1] \rightarrow [0,1]$ using the detections with higher scores than $\bar{u}^c$.
Still, as $\zeta^c(\cdot)$ changes the confidence scores, we need another threshold $\bar{v}^c$, as the operating threshold, to remove the redundant detections after calibration.
Following the accuracy measure, we cross-validate $\bar{u}^c$ and $\bar{v}^c$ using \gls{LRP}.
As for inference time, for the $i$-th detection $(\hat{p}_i, \hat{b}_i, \hat{c}_i)$, if $\hat{p}_i \geq \bar{u}^{\hat{c}_i}$, it survives to the calibrator and then $\hat{p}^{cal}_i=\zeta^{\hat{c}_i}(\hat{p}_i)$.
Finally, if $\hat{p}^{cal}_i \geq \bar{v}^{\hat{c}_i}$, the $i$-th detection is an output of the detector.
Please see Alg. \ref{alg:training} and \ref{alg:inference} for the details of training and inference.
We now describe the specific models for $\zeta^c(\cdot)$ and how we optimize them. 
Overall, we prefer monotonically increasing functions as $\zeta^c(\cdot)$ in order not to affect the ranking of the detections significantly and to keep their accuracy. 

\blockcomment{
\textbf{Calibration Target for \gls{ELCE}}
\begin{align}
\label{eq:ELCEbin}
  &\Bigg\lvert \sum_{\substack{\hat{b}_i \in \hat{\mathcal{D}}_j^c}} \big( \hat{p}_i - \mathrm{lq}(\hat{b}_i, b_{\psi(i)}) \big)  \Bigg\rvert.
\end{align}
}

\textbf{Distribution Calibration via Platt Scaling} Assuming that $\hat{p}_i$ is sampled from Bernoulli distribution $\mathcal{B}(\cdot)$, we aim to minimize the \gls{NLL} of the predictions on the target distribution $\mathcal{B}(\mathrm{IoU}(\hat{b}_i, b_{\psi(i)}))$ using \gls{PS} \cite{plat2000probabilistic}.
Accordingly, we recover the logits, and then shift and scale the logits to obtain the calibrated probabilities $\hat{p}^{cal}_i$,
\begin{align}
    \label{eq:ps}
    \hat{p}^{cal}_i = \sigma ( a\sigma^{-1}(\hat{p}_i) + b),
\end{align}
where $\sigma(\cdot)$ is the sigmoid and $\sigma^{-1}(\cdot)$ is its inverse, as well as $a \geq 0$ and $b$ are the learnable parameters.
We derive the \gls{NLL} for the $i$th detection in App. \ref{app:method} as
\begin{align}
\label{eq:nll}
  - (\mathrm{IoU}(\hat{b}_i, b_{\psi(i)}) \log(\hat{p}^{cal}_i) + (1-\mathrm{IoU}(\hat{b}_i, b_{\psi(i)})) \log(1-\hat{p}^{cal}_i)).
\end{align}
Please note that Eq. \eqref{eq:nll}, which is in fact the cross-entropy loss, is minimized if $\hat{p}^{cal}_i=\mathrm{IoU}(\hat{b}_i, b_{\psi(i)})$ when $\mathrm{LaECE}_0$ and $\mathrm{LaACE}_0$ are minimized.
We optimize Eq. \eqref{eq:nll} via the second-order optimization strategy L-BFGS \cite{L-BFGS} following \cite{CalibrationOD}.

\textbf{Confidence Calibration via Isotonic Regression}
As an alternative perspective, $\hat{p}_i$ can also be directly calibrated by modelling the calibration as a regression task.
To do so, we construct the prediction-target pairs ($\{\hat{p}_i, \mathrm{IoU}(\hat{b}_i, b_{\psi(i)})\}$) on the held-out val. set and then fit an \gls{IR} model using scikit-learn \cite{scikit-learn}.

\textbf{Adapting Our Approach to Different Calibration Objectives} 
Until now, we considered post-hoc calibrators for $\mathrm{LaECE}_0$ and $\mathrm{LaACE}_0$ though in practice different measures can be preferred.
Our post-hoc calibrators can easily be adapted for such cases by considering the dataset design and optimisation criterion.
To illustrate, for \gls{DECE}-style evaluation, the calibration dataset is to be class-agnostic where the detections are thresholded from $0.30$ with prediction-target pairs for \gls{IR} as $(\{\hat{p}_i, 0\})$ and $(\{\hat{p}_i, 1\})$ for \glspl{FP} and  \glspl{TP} respectively.
\section{Experimental Evaluation}
\label{sec:experiments}

We now show that our post-hoc calibration approaches consistently outperform training time calibration methods by significant margins (\cref{subsec:comparison_sota}) and that they generalize to any detector and can thus be used as a strong baseline (\cref{subsec:differentmethods}).

\subsection{Comparing Our Baselines with \gls{SOTA} Calibration Methods}
\label{subsec:comparison_sota}
Here, we compare \gls{PS} and \gls{IR} with recent training-time calibration methods considering various evaluation approaches.
As these training-time methods mostly rely on D-DETR, we also use D-DETR with ResNet-50 \cite{ResNet}.
We obtain the detectors of training time approaches trained with COCO dataset from their official repositories, whereas we incorporate Cityscapes into their official repositories and train them using the recommended setting in \cref{tab:proper_training_cityscapes}. 

\begin{table}[t]
\small
\setlength{\tabcolsep}{1.0em}
\centering
\caption{Comparison with \gls{SOTA} methods in terms of other evaluation measures on COCO \cite{COCO}. LRP is reported on LRP-optimal thresholds obtained on val. set. AP is reported on top-100 detections. $\tau$ is taken as $0.50$. All measures are lower-better, except AP. \textbf{Bold:} the best, \underline{underlined}: second best. \gls{PS}: Platt Scaling, \gls{IR}: Isotonic Regression. 
}
\label{tab:existing_eval_coco}
\scalebox{1}{
\begin{tabular}{c|c||c|c||c|c||c|c} 
\toprule
\midrule
Cal.&&\multicolumn{2}{c||}{Calibration (thr. 0.30)}&\multicolumn{2}{c||}{Calibration (LRP thr.)}&\multicolumn{2}{c}{Accuracy}\\ 
%
%
Type&Method&\gls{DECE}&$\mathrm{LaECE}$&\gls{DECE}&$\mathrm{LaECE}$&$\mathrm{LRP}$&$\mathrm{AP}\uparrow$ \\ \midrule
Uncal.&D-DETR \cite{DDETR}
&$12.8$&$13.2$
&$15.0$&$12.1$
&$66.3$&$44.1$\\
\midrule
\multirow{5}{*}{Training}
&MbLS \cite{liu2023MbLS}
&$15.6$&$16.3$
&$18.7$&$15.8$
&$65.9$&$44.3$\\
&MDCA \cite{hebbalaguppe2022MDCA}
&$12.2$&$13.5$
&$14.3$&$12.6$
&$66.4$&$43.8$\\
&TCD \cite{munir2022tcd}
&$12.4$&$13.1$
&$14.4$&$12.3$
&$66.6$&$44.0$\\
Time&BPC \cite{munir2023bpc}
&$9.8$&$13.1$
&$11.4$&$12.3$
&$66.8$&$43.6$\\
&Cal-DETR \cite{munir2023caldetr}
&$8.7$&$12.9$
&$9.7$&$11.8$
&$66.0$ &$44.4$\\
\midrule
&\gls{PS} for \gls{DECE}
&$\mathbf{0.9}\imp{7.8}$&$16.3$
&$\mathbf{2.4}\imp{7.3}$&$15.8$
&$66.3$&$44.1$\\
Post-hoc&\gls{PS} for \gls{LaECE}
&$11.0$&$\underline{11.5}\imp{1.4}$
&$9.4$&$\underline{10.1}\imp{1.7}$
&$66.3$&$44.1$\\
(Ours)&\gls{IR} for \gls{DECE}
&$\underline{1.3}\imp{7.4}$&$15.7$
&$\underline{2.6}\imp{7.1}$&$15.3$
&$66.2$&$44.1$\\
&\gls{IR} for \gls{LaECE}
&$10.2$&$\textbf{8.9}\imp{4.0}$
&$9.3$&$\textbf{8.2}\imp{3.6}$
&$66.3$&$43.7$\\
\midrule
\bottomrule
\end{tabular}}
\end{table}
\textbf{Comparison on Other Evaluation Approaches} 
Before moving on to our evaluation approach, we first show that our \gls{PS} and \gls{IR} outperform all existing training time methods on existing evaluation approaches.
For that, we consider \gls{DECE} and the \gls{LaECE} from $\tau=0.5$ by including two different evaluation settings for each: (i) the detection set is obtained from the fixed threshold of $0.30$ following the convention \cite{munir2023caldetr,CalibrationOD,munir2022tcd,munir2023bpc}, and (ii) the operating thresholds are cross-validated using \gls{LRP}.
Following their standard usage, we use 10  and 25 bins to compute \gls{DECE} and \gls{LaECE} respectively.
We optimize \gls{PS} and \gls{IR} by considering the calibration objective as described in \cref{sec:posthoc}.
\cref{tab:existing_eval_coco} shows that \gls{PS} and \gls{IR} outperform \gls{SOTA} Cal-DETR significantly by more than $7$ \gls{DECE} and up to $4$ \gls{LaECE} on COCO \textit{minitest}.
Please note that \textit{all previous approaches are optimized for \gls{DECE} thresholded from $0.30$, in terms of which our \gls{PS} yields only $0.9$ \gls{DECE} improving the \gls{SOTA} by $7.8$.}
Finally, \cref{tab:existing_eval_coco} suggests that post-hoc calibrators perform the best when the calibration objective is aligned with the measure.
App. \ref{app:experiments} shows that our observations also generalize to Cityscapes.

\begin{table}[t]
\small
\setlength{\tabcolsep}{0.2em}
\centering
\caption{Comparison with \gls{SOTA} calibration methods on Common Objects using our proposed evaluation. Our gains (green/red) are reported for \gls{IR} compared to the best existing approach. \textbf{Bold:} the best, \underline{underlined}: second best in terms of calibration.}
\label{tab:coco}
\scalebox{1}{
\begin{tabular}{c|c||c|c|c||c|c|c||c|c|c} 
\toprule
\midrule
&&\multicolumn{3}{c||}{COCO \textit{minitest}}&\multicolumn{3}{c||}{COCO-C}&\multicolumn{3}{c}{Obj45K}\\ 
Calibration&&\multicolumn{3}{c||}{(\gls{ID})}&\multicolumn{3}{c||}{(Domain Shift)}&\multicolumn{3}{c}{(Domain Shift)}\\ 
%
%
Type&Method&$\mathrm{\gls{LaECE}}_{0}$&$\mathrm{\gls{LaACE}}_{0}$&$\mathrm{LRP}$&$\mathrm{\gls{LaECE}}_{0}$&$\mathrm{\gls{LaACE}}_{0}$&$\mathrm{LRP}$&$\mathrm{\gls{LaECE}}_{0}$&$\mathrm{\gls{LaACE}}_{0}$&$\mathrm{LRP}$ \\ \midrule
Uncalibrated&D-DETR \cite{DDETR} &$12.7$&$27.1$&$57.3$
&$14.6$&$28.7$&$71.5$
&$\underline{16.4}$& $35.8$ & $72.0$\\
\midrule
\multirow{5}{*}{Training-time}
&MbLS \cite{liu2023MbLS}&$16.5$&$30.3$&$56.8$
&$16.8$&$31.1$&$71.8$
&$17.3$&$37.1$&$71.6$ \\
&MDCA \cite{hebbalaguppe2022MDCA} &$13.1$&$27.2$&$57.5$
&$14.5$&$28.7$&$71.8$
&$16.6$&$35.6$&$72.2$\\
&TCD \cite{munir2022tcd} & $13.0$& $26.7$& $57.6$
&$14.6$&$28.3$&$71.9$
&$\mathbf{16.3}$& $35.5$& $71.7$ \\
&BPC \cite{munir2023bpc}&$12.4$&$25.5$ & $57.7$
&$14.1$&$27.1$&$72.1$
&$17.3$&$34.5$& $72.0$\\
&Cal-DETR \cite{munir2023caldetr} &$11.6$&$24.6$&$56.2$
&$13.8$&$26.4$&$70.6$
&$18.8$& $35.3$ & $71.1$\\
\midrule
Post-hoc&Platt Scaling &$\underline{9.6}$ & $\underline{23.5}$& $57.3$
&$\underline{12.8}$&$\underline{25.6}$&$71.5$
&$17.0$&$\underline{33.7}$&$72.0$\\
(Ours)&Isotonic Regression  & $\mathbf{7.7}$& $\mathbf{23.1}$& $57.2$
&$\mathbf{10.7}$&$\mathbf{25.3}$&$71.5$
&$17.2$&$\textbf{33.3}$&$72.0$\\
&&\imp{3.9}&\imp{1.5}& &\imp{3.1}&\imp{1.1}&&\nimp{0.9}&\imp{1.2}&\\ 
\midrule
\bottomrule
\end{tabular}}
\end{table}

\begin{table}[t]
\small
\setlength{\tabcolsep}{0.2em}
\centering
\caption{Comparison with \gls{SOTA} on Autonomous Driving using our proposed evaluation. Our gains (green/red) are reported for \gls{IR} compared to the best existing approach. \textbf{Bold:} the best, \underline{underlined}: second best in terms of calibration.}
\label{tab:cs}
\scalebox{1.00}{
\begin{tabular}{c|c||c|c|c||c|c|c||c|c|c} 
\toprule
\midrule
&&\multicolumn{3}{c||}{Cityscapes \textit{minitest}}&\multicolumn{3}{c||}{Cityscapes-C}&\multicolumn{3}{c}{Foggy Cityscapes}\\ 
Calibration&&\multicolumn{3}{c||}{(\gls{ID})}&\multicolumn{3}{c||}{(Domain Shift)}&\multicolumn{3}{c}{(Domain Shift)}\\ 
%
%
Type&Method&$\mathrm{\gls{LaECE}}_{0}$&$\mathrm{\gls{LaACE}}_{0}$&$\mathrm{LRP}$&$\mathrm{\gls{LaECE}}_{0}$&$\mathrm{\gls{LaACE}}_{0}$&$\mathrm{LRP}$&$\mathrm{\gls{LaECE}}_{0}$&$\mathrm{\gls{LaACE}}_{0}$&$\mathrm{LRP}$ \\ \midrule
Uncalibrated&D-DETR \cite{DDETR} &$20.3$&$26.0$&$57.2$
&$21.4$&$\textbf{25.6}$&$80.2$
&$18.5$&$22.3$&$69.4$ \\
\midrule
\multirow{3}{*}{Training-time}
&TCD \cite{munir2022tcd} &$16.8$&$31.7$&$59.2$
&$23.2$&$32.4$&$81.6$
&$24.4$&$33.8$&$71.6$ \\
&BPC \cite{munir2023bpc} &$23.8$&$31.8$&$64.9$
&$28.1$&$33.3$&$83.7$
&$24.7$&$30.9$&$73.8$ \\ 
&Cal-DETR \cite{munir2023caldetr} &$21.3$&$25.3$&$56.9$
&$23.0$&$26.4$&$80.8$
&$20.0$&$23.2$&$71.0$ \\
\midrule
Post-hoc&Platt Scaling&$\underline{9.6}$&$\mathbf{23.3}$&$57.2$
&$\underline{17.7}$&$26.2$&$80.2$
&$\underline{11.3}$&$\underline{21.6}$&$69.4$ \\
(Ours)&Isotonic Regression&$\mathbf{9.0}$&$\underline{23.7}$&$56.8$
&$\textbf{16.4}$&$\underline{25.8}$&$80.5$
&$\textbf{10.0}$&$\textbf{21.2}$&$69.5$ \\
&&\imp{7.8}&\imp{1.6}& &\imp{5.0}&\nimp{0.2}&&\imp{8.5}&\imp{1.1}&\\ 
\midrule
\bottomrule
\end{tabular}}
\end{table}
\textbf{Common Objects Setting} 
We now evaluate detectors using our evaluation approach.
\cref{tab:coco} shows that \gls{IR} and \gls{PS} share the top-2 entries on almost all test subsets by preserving the accuracy (\gls{LRP}) of D-DETR.
Specifically, our gains on \gls{ID} set and COCO-C are significant, where \gls{IR} outperforms Cal-DETR by around $3-4 $ $\mathrm{\gls{LaECE}}_{0}$ and $1.0-1.5$ $\mathrm{\gls{LaACE}}_{0}$.
As for Obj45K, the challenging test set with natural shift, \gls{IR} and \gls{PS} improve $\mathrm{\gls{LaACE}}_{0}$ but perform slightly worse in terms of $\mathrm{\gls{LaECE}}_{0}$.
This is an expected drawback of post-hoc approaches when the domain shift is large as they are trained only with \gls{ID} val. set \cite{domainshiftposthoc}.

\textbf{Autonomous Driving Setting}
\cref{tab:cs} shows that our approaches consistently outperform all training time calibration approaches on this setting as well.
Specifically, our gains are very significant ranging between $5.0$-$8.5$ $\mathrm{\gls{LaECE}}_{0}$ compared to the \gls{SOTA} Cal-DETR, further presenting the efficacy of our approaches.

\begin{table}[t]
\small
\setlength{\tabcolsep}{0.1em}
\centering
\caption{Comparison with \gls{TS} using D-DETR. \textbf{Bold:} the best, \underline{underlined}: second best. \xmark\ : domain-shifted val. set is used to obtain thresholds and calibrators, decreasing the accuracy (\textcolor{red}{red} font). Bias term only exists for \gls{PS} ($b$ in Eq. \eqref{eq:ps}), thus N/A for \gls{IR}.}
\vspace{-2.ex}
\label{tab:ablation}
\scalebox{0.95}{
\begin{tabular}{c||c|c||c|c||c|c|c||c|c|c} 
\toprule
\midrule
%
%
&\multicolumn{2}{c||}{Ablations on Dataset}&\multicolumn{2}{c||}{Ablations on Model} & \multicolumn{3}{c||}{COCO \textit{minitest}} & \multicolumn{3}{c}{Cityscapes \textit{minitest}}\\%

Method&ID Val. Set&Threshold&Class-wise&Bias Term&$\mathrm{LaECE}_{0}$&$\mathrm{LaACE}_{0}$&$\mathrm{LRP}$ &$\mathrm{LaECE}_{0}$&$\mathrm{LaACE}_{0}$&$\mathrm{LRP}$ \\ \midrule
Temperature Scaling&\xmark& & &   & $12.3$& $\mathbf{20.8}$ & \textcolor{red}{$61.5$} & $21.0$&$25.9$&\textcolor{red}{$60.3$} \\ 
(Current Baseline)&\cmark& &&& $12.5$ & $23.1$& $57.3$ & $20.9$ & $26.3$&$57.2$ \\ \midrule
Ablations on&\cmark&\cmark&& & $11.3$ & $24.8$ & $57.3$ & $13.3$ &$25.5$&$57.2$ \\
Temperature &\cmark& & \cmark& & $12.4$ & $\underline{22.9}$& $57.3$ & $23.1$&$27.5$&$57.2$ \\
Scaling&\cmark&\cmark&\cmark & & $10.6$ &$24.2$ & $57.3$ & $12.7$&$24.6$&$57.2$ \\
%
\midrule
Platt Scaling (Ours)&\cmark&\cmark&\cmark&\cmark& $\underline{9.6}$& $23.5$ & $57.3$ & $\underline{9.6}$ &$\mathbf{23.3}$&$57.2$ \\

Isotonic Regression (Ours)&\cmark&\cmark&\cmark &N/A&$\textbf{7.7}$& $23.1$ & $57.2$  & $\textbf{9.0}$&$\underline{23.7}$&$56.8$ \\
\bottomrule
\end{tabular}}
\end{table}

\textbf{Comparison with Existing Temperature Scaling Baseline and Ablations}
\cref{tab:ablation} compares \gls{TS} for different design choices as well as with our \gls{PS} and \gls{IR}.
Please note that \xmark \  corresponds to the baseline setting used in the recent approaches \cite{munir2022tcd, MCCL, munir2023bpc, munir2023caldetr} that employ Objects365 \cite{Objects365} and BDD100K \cite{Objects365} as domain-shifted val. sets for obtaining the calibrator.
Due to this domain shift, the accuracy of \gls{TS} degrades by up to $4$ \gls{LRP}, in red font, as the operating thresholds obtained on these val. sets do not generalize to the \gls{ID} set; showing that it is crucial to use an \gls{ID} val set.
In ablations, thresholding the detections and class-wise calibrators generally improves the performance of \gls{TS} and a more notable gain is observed once the bias term is used in \gls{PS}.
\textit{Our \gls{PS} outperforms \gls{TS} baseline obtained on \gls{ID} val. set by $\sim 3$  $\mathrm{LaECE}_{0}$ on COCO and $11.4$ $\mathrm{LaECE}_{0}$  on Cityscapes.}
Finally, \gls{IR} performs on par or better compared to \gls{PS}.

\subsection{Calibrating and Evaluating Different Detection Methods} 
\label{subsec:differentmethods}

\begin{table}[t]
\small
\setlength{\tabcolsep}{0.1em}
\centering
\caption{Calibrating and evaluating different object detectors. We use Common Objects setting and report the results on COCO \textit{minitest}. $^*$ denotes the detectors in \cref{fig:question}. Among these detectors, considering the uncalibrated columns, our evaluation ranks the D-DETR as the best. \textbf{Bold:} the best, \underline{underlined}: second best for calibration.}
\label{tab:model_zoo_coco}
\scalebox{0.95}{
\begin{tabular}{c|c|c|c|c|c||c|c|c||c|c|c||c} 
\toprule
\midrule
&&&\multicolumn{3}{c||}{Uncalibrated}&\multicolumn{3}{c||}{Platt Scaling}&\multicolumn{3}{c||}{Isotonic Regression}&\\ 
%
%
Type&Detector&Backbone&$\mathrm{\gls{LaECE}}_{0}$&$\mathrm{\gls{LaACE}}_{0}$&$\mathrm{LRP}$&$\mathrm{\gls{LaECE}}_{0}$&$\mathrm{\gls{LaACE}}_{0}$&$\mathrm{LRP}$&$\mathrm{\gls{LaECE}}_{0}$&$\mathrm{\gls{LaACE}}_{0}$&$\mathrm{LRP}$&$\mathrm{AP}\uparrow$ \\ \midrule
\multirow{5}{*}{One-Stage}
&PAA \cite{paa}$^*$&R50
&$15.9$&$28.1$&$59.7$
&\underline{$9.7$}&\underline{$24.3$}&$59.7$
&$\mathbf{7.7}$&$\mathbf{23.8}$&$59.7$
&$43.2$\\
&ATSS \cite{ATSS}$^*$&R50
&$19.1$&$34.0$&$59.5$
&\underline{$10.3$}&\underline{$24.7$}&$59.5$
&$\mathbf{8.5}$&$\mathbf{24.1}$&$59.5$
&$43.1$\\
&GFL \cite{GFL}&R50
&$13.7$&$28.5$&$59.3$
&\underline{$10.3$}&\underline{$24.5$}&$59.3$
&$\mathbf{8.3}$&$\mathbf{24.0}$&$59.3$
&$43.0$\\
&VFNet \cite{varifocalnet}&R50
&$13.9$&$25.8$&$57.7$
&\underline{$10.7$}&\underline{$25.1$}&$57.7$
&$\mathbf{8.3}$&$\mathbf{24.6}$&$57.7$
&$44.8$\\
\midrule
\multirow{2}{*}{Two-Stage}
&Faster R-CNN\cite{FasterRCNN}$^*$&R50
&$27.0$&$29.9$&$60.4$
&\underline{$10.4$}&\underline{$23.8$}&$60.4$
&$\mathbf{8.6}$&$\mathbf{23.5}$&$60.4$
&$40.1$\\
&RS R-CNN\cite{RSLoss}$^*$&R50
&$19.7$&$28.9$&$58.7$
&\underline{$10.2$}&\underline{$23.5$}&$58.7$
&$\mathbf{8.1}$&$\mathbf{23.0}$&$58.8$
&$42.4$\\
\midrule
\multirow{4}{*}{DETR-like}
&D-DETR \cite{DDETR}$^*$& R50
&$12.7$&$27.1$&$57.3$
&$\underline{9.6}$&\underline{$23.5$}&$57.3$
&$\mathbf{7.7}$&$\mathbf{23.1}$&$57.2$
&$44.1$\\
&UP-DETR \cite{dai2022updetr}& R50
&$34.4$&$35.2$&$55.8$
&\underline{$10.0$}&$\underline{22.6}$&$55.8$
&$\mathbf{8.2}$&$\mathbf{22.2}$&$55.9$
&$42.9$\\
&DINO\cite{zhang2022dino}&R50
&$13.6$&$26.9$&$53.6$
&\underline{$10.6$}&\underline{$23.5$}&$53.6$
&$\mathbf{8.9}$&$\mathbf{22.8}$&$53.6$
&$50.4$\\
\midrule
\multirow{2}{*}{OVOD}
&GLIP \cite{glip}&Swin-T
&$13.0$&$25.3$&$49.0$
&\underline{$9.2$}&\underline{$22.4$}&$49.0$
&$\mathbf{7.7}$&$\mathbf{21.8}$&$49.0$
&$55.7$\\
&G. DINO \cite{groundingdino}&Swin-T
&$13.8$&$27.5$&$46.9$
&$\underline{8.9}$&$\underline{21.9}$&$46.9$
&$\mathbf{7.7}$&$\mathbf{21.3}$&$47.0$
&$58.3$\\
\midrule
\multirow{3}{*}{SOTA}
&Co-DETR \cite{codetr}&Swin-L
&$10.8$&$23.0$&$41.5$
&\underline{$8.6$}&\underline{$20.2$}&$41.5$
&$\mathbf{6.7}$&$\mathbf{19.3}$&$41.6$
&$64.5$\\
&EVA \cite{EVA}&ViT(EVA)
&$17.4$&$21.2$&$41.2$
&\underline{$8.6$}&\underline{$20.3$}&$41.2$
&$\mathbf{7.1}$&$\mathbf{19.9}$&$41.2$
&$64.5$\\
&MoCaE \cite{MOCAE}&N/A
&$10.6$&$21.4$&$40.7$
&\underline{$8.9$}&\underline{$20.4$}&$40.7$
&$\mathbf{7.3}$&$\mathbf{19.9}$&$40.7$
&$65.0$\\
\midrule
\bottomrule
\end{tabular}}
\end{table}
This section shows another benefit of post-hoc calibration approaches, that is, they generalize to \textit{any} object detector, thus they can be reliably used as baselines.

\begin{wrapfigure}{r}{10cm}
        \captionsetup[subfigure]{}
        \centering
        \begin{subfigure}[b]{0.25\textwidth}
        \centering
            \includegraphics[width=\textwidth]{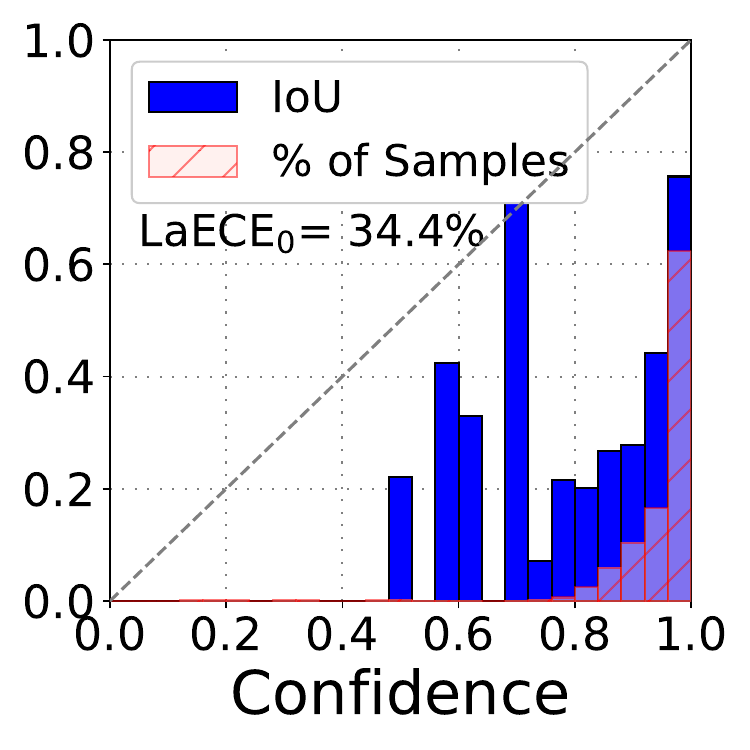}
            \caption{Uncalibrated }
        \end{subfigure}
        \begin{subfigure}[b]{0.25\textwidth}
        \centering
            \includegraphics[width=\textwidth]{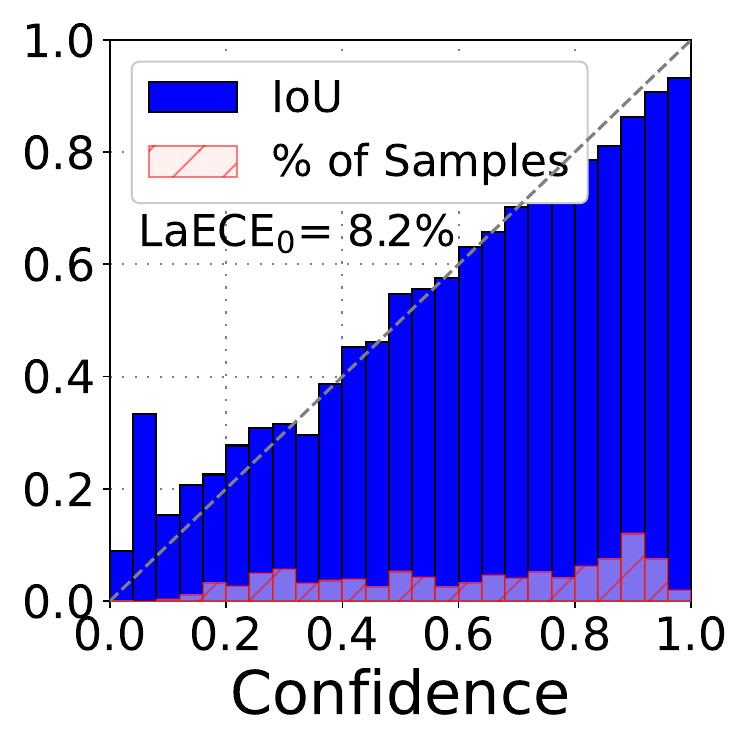}
            \caption{Calibrated by \gls{IR}}
        \end{subfigure}
        \caption{The reliability diagrams of UP-DETR.
        }
        \vspace{-2.ex}
        \label{fig:up_detr_reliability}
\end{wrapfigure}

\textbf{Calibrating Any Detector with Post-hoc Approaches} In \cref{tab:model_zoo_coco}, we calibrate 14 different detectors with a very diverse set of architectures 
using \gls{PS} and \gls{IR}.
The results suggest that \gls{IR} consistently performs better than \gls{PS}, which outperforms an uncalibrated detector.
Specifically, \gls{IR} decreases the range of $\mathrm{\gls{LaECE}}_{0}$ from $10.6-34.4$ to $6.7-8.9$ on COCO \textit{minitest} by preserving the accuracy, making it a solid baseline.
As an example, \cref{fig:up_detr_reliability} provides the reliability diagrams of the overconfident UP-DETR and shows that \gls{IR} significantly improves its calibration quality.
We also use EVA, Co-DETR and MoCaE, the combination of these two as a mixture of experts, as the most accurate publicly-available detectors on COCO dataset.
Our results show that MoCaE performs the best in terms of accuracy with $40.7$ \gls{LRP} while Co-DETR has the best calibration with $6.7$ $\mathrm{\gls{LaECE}}_{0}$ and $19.3$ $\mathrm{\gls{LaACE}}_{0}$, which the future work should aim to surpass.

\textbf{Calibration Under Long-tailed Class Distribution} 
As common baselines used for LVIS dataset, here we use Mask R-CNN \cite{MaskRCNN} and Cascade Mask R-CNN \cite{CascadeRCNN} along with their stronger versions trained with Seesaw Loss \cite{seesawloss}.
We obtain two calibrators for each class using the held-out LVIS \textit{minival}: (i) for object detection using the \gls{IoU}; and (ii) for instance segmentation using mask \gls{IoU} as the calibration target.
\cref{tab:lvis} shows that \gls{IR} improves calibration by up to $9$ $\mathrm{\gls{LaECE}}_{0}$ and $2.8$ $\mathrm{\gls{LaACE}}_{0}$, showing that it is still a strong baseline for this challenging setting with around 1K classes.
However, $\mathrm{\gls{LaECE}}_{0}$ and $\mathrm{\gls{LaACE}}_{0}$ are greater compared to COCO (\cref{tab:coco}) and Cityscapes (\cref{tab:cs}) \textit{minitest}, suggesting the need for further research on calibration under long-tailed data.

\textbf{Interpretability of our Evaluation Framework} Finally, we provide an example on interpreting the performance measures of our evaluation framework.
For accuracy, \gls{LRP} is a weighted combination of its \gls{FP}, \gls{FN} and localisation error components (App. \ref{app:related}), which, as an example, are $10.1$, $22.7$ and $18.8$ respectively for Co-DETR \cite{codetr} calibrated with \gls{IR}.
Also considering $\mathrm{\gls{LaACE}}_{0}=19.3$, one can easily infer that: \textit{once deployed with the operating thresholds determined by our framework, Co-DETR finds $77.3$\% of the objects with $89.9$\% precision and $81.2$\% \gls{IoU} where $19.3$\% is the mean absolute error of the  confidence to represent \gls{IoU}}.
We believe these intuitive measures will enable practicioners to make better decision when deploying object detectors.

\begin{table}[t]
\small
\setlength{\tabcolsep}{0.15em}
\centering
\caption{Calibrating and evaluating instance segmentation methods on Long-tailed Objects setting based on LVIS. Results are reported on LVIS minitest, please refer to App. \ref{app:experiments} for the domain-shifted LVIS minitest-C. \textbf{Bold:} the best calibration approach.}
\label{tab:lvis}
\scalebox{0.82}{
\begin{tabular}{c|c|c|c||c|c|c||c||c|c|c||c|c|c||c} 
\toprule
\midrule
&\multicolumn{7}{c||}{\underline{Object Detection}}&\multicolumn{7}{c}{\underline{Instance Segmentation}}\\
&\multicolumn{3}{c||}{Uncalibrated}&\multicolumn{3}{c||}{Isotonic Regression}&Box&\multicolumn{3}{c||}{Uncalibrated}&\multicolumn{3}{c||}{Isotonic Regression}&Mask\\ 
%
%
Detector&$\mathrm{\gls{LaECE}}_{0}$&$\mathrm{\gls{LaACE}}_{0}$&$\mathrm{LRP}$&$\mathrm{\gls{LaECE}}_{0}$&$\mathrm{\gls{LaACE}}_{0}$&$\mathrm{LRP}$&$\mathrm{AP}\uparrow$&$\mathrm{\gls{LaECE}}_{0}$&$\mathrm{\gls{LaACE}}_{0}$&$\mathrm{LRP}$&$\mathrm{\gls{LaECE}}_{0}$&$\mathrm{\gls{LaACE}}_{0}$&$\mathrm{LRP}$&$\mathrm{AP}\uparrow$ \\ \midrule
Mask R-CNN \cite{MaskRCNN}
&$25.2$&$30.4$&$74.7$
&$\mathbf{17.1}$&$\mathbf{28.0}$&$74.6$
&$27.1$
&$25.5$&$30.6$&$75.3$
&$\mathbf{17.6}$&$\mathbf{28.4}$&$75.3$
&$25.9$\\
Seesaw Mask R-CNN \cite{seesawloss}
&$25.0$&$30.2$&$73.1$
&$\mathbf{16.8}$&$\mathbf{27.8}$&$73.0$
&$31.8$
&$25.0$&$30.2$&$73.7$
&$\mathbf{16.7}$&$\mathbf{27.6}$&$73.7$
&$31.0$\\
Seesaw Cascade R-CNN \cite{seesawloss}
&$26.4$&$31.5$&$70.7$
&$\mathbf{17.4}$&$\mathbf{28.7}$&$70.8$
&$36.0$
&$25.5$&$30.6$&$71.7$
&$\mathbf{17.2}$&$\mathbf{28.1}$&$71.6$
&$33.1$\\
\midrule
\bottomrule
\end{tabular}}
\end{table}

\section{Conclusions}
The progress in a field heavily relies on the evaluation tools and the baselines used.
In this paper, we showed that existing evaluation tools for calibration as well as the baseline post-hoc calibrators for object detectors have significant drawbacks.
We remedied that by introducing an evaluation framework including baseline post-hoc calibrators tailored to object detection.
Our experiments suggested that, once evaluated and designed properly, the post-hoc calibrators significantly outperform all existing training-time calibrators.
This implies the need for research to develop better calibration techniques for object detection, for which, we believe, our evaluation framework will be an essential pillar.

\bibliographystyle{splncs04}
\bibliography{references}
\clearpage

\section*{APPENDICES}
\renewcommand{\thefigure}{A.\arabic{figure}}
\renewcommand{\thetable}{A.\arabic{table}}
\renewcommand{\theequation}{A.\arabic{equation}}
\renewcommand{\thealgorithm}{A.\arabic{algorithm}}

\renewcommand{\thesection}{A}
\section{Further Details on Related Work}
\label{app:related}

\textbf{Calibration Error in \cite{popordanoska2024CE}}
Popordanoska et al. \cite{popordanoska2024CE} recently proposed Calibration Error that generalise both \gls{DECE} and \gls{LaECE} through a link function $\mathrm{L}(\hat{b}_i, b_{\psi(i)})$ that can be considered as a generic accuracy term.
In particular, 
$\mathrm{L}(\hat{b}_i, b_{\psi(i)})$ measures the similarity between a detection box $\hat{b}_i$ and its corresponding ground truth bounding box $b_{\psi(i)}$.
Then, following the conventional calibration literature, the predicted confidence is aimed to be aligned with this link function as the accuracy.
\blockcomment{
This can be expressed as the following calibration criterion:
\begin{align}
    \mathbb{E}_{\hat{b}_i \in B_i(\hat{p}_i)}[\mathrm{L}(\hat{b}_i, b_{\psi(i)})] = \hat{p}_i, \forall \hat{p}_i \in [0,1],
\end{align}
}
Owing to the generic definition of accuracy, it is thus possible recover different calibration error measures with this notion.
Specifically, if the link function $\mathrm{L}(\hat{b}_i, b_{\psi(i)})$ is an indicator function that returns true if the detection is a \gls{TP} and has an \glspl{IoU} of at least $\tau$, then \gls{DECE} can be recovered.
A similar recovery holds for \gls{LaECE}, when the link function is taken as the \gls{IoU}, that is $\mathrm{L}(\hat{b}_i, b_{\psi(i)})=\mathrm{IoU}(\hat{b}_i, b_{\psi(i)})$.
Instead of the classical binning approach used in \gls{DECE} and \gls{LaECE}, the authors define calibration error by utilising a kernel  $\mathrm{k}(p_{i}, p_{j})$ to approximate the bins as follows:
\begin{align}
\label{eq:ce}
   \hat{CE} 
    = \frac{1}{K} \sum_{c=1}^{K} \sum_{i \in \hat{\mathcal{D}}^{c}} \frac{1}{|\hat{\mathcal{D}}^{c}|} \left\lvert \hat{p}_i - \frac{\sum_{j \in \hat{\mathcal{D}}, i \neq j}\mathrm{k}(\hat{p}_{i}, \hat{p}_{j}) \mathrm{L}(\hat{b}_i, b_{\psi(i)})}{\sum_{j \in \hat{\mathcal{D}}, i \neq j}\mathrm{k}(\hat{p}_{i}, \hat{p}_{j})} \right\rvert,
\end{align}
where the class-wise errors are averaged over to obtain the final calibration performance.

\textbf{The Components of \gls{LRP} Error} In Eq. \eqref{eq:LRPdefcompact} of \cref{sec:relatedwork}, we defined \gls{LRP} Error. While that definition is intuitive as it combines all three types of errors, i.e., precision, recall and localisation errors, into a single measure, these types of errors are not quantified precisely.
To address this and provide insight on the detector, Oksuz et al. \cite{LRPPAMI} showed that Eq. \eqref{eq:LRPdefcompact} can be rewritten alternatively in the following form:%
\begin{align}
\label{eq:LRPdef}
\mathrm{LRP} = \frac{1}{\mathrm{N_{FP}} +\mathrm{N_{FN}}+\mathrm{N_{TP}}} \left( \mathrm{w_{Loc}} \mathrm{LRP_{Loc}}+ \mathrm{w_{FP}} \mathrm{LRP_{FP}} + \mathrm{w_{FN}} \mathrm{LRP_{FN}} \right),
\end{align}
with the weights $w_{Loc}=\mathrm{N_{TP}}$, $w_{FP}=|\mathcal{D}|$, and $w_{FN}=|\mathcal{G}|$ controlling the contributions of each error type\footnote{Following our design choice in our evaluation framework, here we use $\tau=0$. Hence,  $\mathcal{E}_{loc}(i)= \frac{1- \mathrm{IoU}(\hat{b}_i, b_{\psi(i)})}{1-\tau}$ in Eq. \eqref{eq:LRPdefcompact} reduces to $\mathcal{E}_{loc}(i)=1- \mathrm{IoU}(\hat{b}_i, b_{\psi(i)})$ and $w_{Loc}=\mathrm{N_{TP}}$.}. 
Then, denoting the set of all detections $\hat{\mathcal{D}}$, $\mathrm{LRP_{Loc}}$ measures the average localisation error of the \glspl{TP} detections ($\psi(i) > 0$),
\begin{align}
\label{eq:Loc}
\mathrm{LRP_{Loc}}=\frac{1}{\mathrm{N_{TP}}}\sum \limits_{i \in \hat{\mathcal{D}}, \psi(i) > 0} \mathcal{E}_{loc}(i).
\end{align}
$\mathrm{LRP_{FP}}$ and $\mathrm{LRP_{FN}}$ correspond to the \gls{FP} and \gls{FN} rates as the precision and recall errors respectively:
\begin{align}
\label{eq:Type1}
\mathrm{LRP_{FP}}=1- \frac{\mathrm{N_{TP}}}{|\hat{\mathcal{D}}|}=\frac{\mathrm{N_{FP}}}{|\hat{\mathcal{D}}|} \text{ and }
\mathrm{LRP_{FN}}=1- \frac{\mathrm{N_{TP}}}{M}=\frac{\mathrm{N_{FP}}}{M},
\end{align}
where $M$ is the number of total objects.
\renewcommand{\thesection}{B}
\section{Analyses of Existing Calibration Measures and Further Details}
\label{app:analyses}

In this section, we first provide further details on \gls{DECE}-style evaluation, which are not included in the main paper due to the space limitation.
Then, we provide our analyses on \gls{LaECE}-style evaluation and \gls{CE}-style evaluation. 

\subsection{Further Details on \gls{DECE}-style Evaluation}
\label{app:further_details_dece_style}

\textbf{Derivation of Eq. \ref{eq:DECEbin}}
First and foremost, Eq. \eqref{eq:dece_} defines \gls{DECE} as:
\begin{align}
\label{eq:dece_app}
   \text{\gls{DECE}} 
    = \sum_{j=1}^{J} \frac{|\hat{\mathcal{D}}_j|}{|\hat{\mathcal{D}}|} \left\lvert \bar{p}_{j} - \mathrm{precision}(j)  \right\rvert,
\end{align}
where $\hat{\mathcal{D}}$ and $\hat{\mathcal{D}}_j$ are the set of all detections and the detections in the $j$-th bin, and $\bar{p}_{j}$ and $\mathrm{precision}(j)$ are the average confidence and the precision of the detections in the $j$-th bin.
With that, precision can be defined as follows:
\begin{align}
 \mathrm{precision}(j) = \frac{\sum_{\hat{b}_i \in \hat{\mathcal{D}}_j, \psi(i) > 0} 1}{|\hat{\mathcal{D}}_j|}.
\end{align}
Therefore, Eq. \eqref{eq:dece_app} can be expressed as:
\begin{align}
\label{eq:dece_app2}
   \text{\gls{DECE}} 
    &= \sum_{j=1}^{J} \frac{|\hat{\mathcal{D}}_j|}{|\hat{\mathcal{D}}|} \left\lvert \frac{\sum_{\hat{b}_i \in \hat{\mathcal{D}}_j} \hat{p}_i}{|\hat{\mathcal{D}}_j|} - \frac{\sum_{\hat{b}_i \in \hat{\mathcal{D}}_j, \psi(i) > 0} 1}{|\hat{\mathcal{D}}_j|}  \right\rvert \\
    &= \frac{1}{|\hat{\mathcal{D}}|}  \sum_{j=1}^{J} \left\lvert \sum_{\hat{b}_i \in \hat{\mathcal{D}}_j} \hat{p}_i - \sum_{\hat{b}_i \in \hat{\mathcal{D}}_j, \psi(i) > 0} 1  \right\rvert 
\end{align}
Decoupling the errors of \glspl{TP} ($\psi(i) > 0$) and \glspl{FP} ($\psi(i) = -1$), and rearranging the terms in the summation, we have
\begin{align}
\label{eq:decenormalizedsum}
    \text{\gls{DECE}}=\frac{1}{|\hat{\mathcal{D}}|}  \sum_{j=1}^{J}  \Bigg\lvert \sum_{\substack{\hat{b}_i \in \hat{\mathcal{D}}_j, \psi(i) > 0}} \big( \hat{p}_i - 1 \big) + \sum_{\substack{\hat{b}_i \in \hat{\mathcal{D}}_j, \psi(i) = -1}} \hat{p}_i \Bigg\rvert.
\end{align}
As a result, \gls{DECE} corresponds to the sum of normalized errors in each bin where the normalization constant is the number of detections ($|\hat{\mathcal{D}}|$), and hence  minimising
\begin{align}
\label{eq:decebinappendix}
\Bigg\lvert \sum_{\substack{\hat{b}_i \in \hat{\mathcal{D}}_j, \psi(i) > 0}} \big( \hat{p}_i - 1 \big) + \sum_{\substack{\hat{b}_i \in \hat{\mathcal{D}}_j, \psi(i) = -1}} \hat{p}_i \Bigg\rvert,
\end{align}
minimises \gls{DECE} for the $j$-th bin, concluding the derivation.

\textbf{Details of \cref{fig:images}} The calibration errors require to match accuracy of a population (a set of detections) with the average confidence score of the same population where the population is commonly represented as the detections in a bin.
That is why, computing a calibration error of the $i$-th detection, as we did in \cref{fig:images}(d-f), requires some assumptions on other detections except the $i$-th one.
In particular, when we plot a calibration error for a single detection, we assume that the confidence scores of all other detections are constant at the point where the calibration error is minimised for. To illustrate this minimisation criterion, \gls{DECE} is minimized when all \glspl{TP} have a confidence of $1$ and \glspl{FP} have a confidence of $0$ as we showed in Eq. \eqref{eq:DECEbin}.  More specifically, 
we assume that the confidence scores of other detections are equal to their target confidences used to obtain the post-hoc calibrators. To make it more clear, we provide an example below.

As an example, when we plot \gls{DECE} for the \gls{TP} detection belonging to the car on the left in \cref{fig:images}(b), we assume that the confidence of the 
 \gls{FP} detection belonging to the car on the right is $0.00$ based on Eq. \eqref{eq:DECEbin}.
 Therefore, the only positive contribution to the \gls{DECE} originates from the detection belonging to the car on the right as the one that we are interested in.
 Now using the alternative (and equivalent) definition of \gls{DECE} we obtained in Eq. \ref{eq:decenormalizedsum} as
\begin{align}
\label{eq:decenormalizedsum_}
    \text{\gls{DECE}}=\frac{1}{|\hat{\mathcal{D}}|}  \sum_{j=1}^{J}  \Bigg\lvert \sum_{\substack{\hat{b}_i \in \hat{\mathcal{D}}_j, \psi(i) > 0}} \big( \hat{p}_i - 1 \big) + \sum_{\substack{\hat{b}_i \in \hat{\mathcal{D}}_j, \psi(i) = -1}} \hat{p}_i \Bigg\rvert,
\end{align}
we only focus on the error originating from a single detection, which is the \gls{TP} car on the right. 
Specifically, (i) ignoring the normalisation constant $|\hat{\mathcal{D}}|$ and (ii) considering that all other detections contribute to \gls{DECE} with an error of $0$ as their confidence matches the target confidence, Eq. \eqref{eq:decenormalizedsum_} reduces to
\begin{align}
    \lvert \hat{p}_i - 1 \rvert=1-\hat{p}_i,
\end{align}
which is the function we plot in \cref{fig:images}(d) for \gls{DECE}.

Similarly, for \gls{DECE}, the calibration error function of a \gls{FP} can be derived as $\hat{p}_i$, which is the function we plot for both (e) and (f).
We can also obtain the \gls{LaECE} of a detection for different measures by following the same methodology, that is $\lvert \hat{p}_i - \mathrm{IoU}(\hat{b}_i, b_{\psi(i)}) \rvert$ for a \gls{TP} (as in (d)) and $\hat{p}_i$ for a \gls{FP} (as in (e) and (f)).
For our measures, as $\tau=0$, we observe that (e) also follows $\lvert \hat{p}_i - \mathrm{IoU}(\hat{b}_i, b_{\psi(i)}) \rvert$. As for the discussion why COCO-style \gls{DECE} remains constant in (e), please refer to App. \ref{sec:CEanalysis}.

\textbf{Training Details of Cityscapes}
Here we present the implementation details of Cityscapes-style training we used to obtain the results in \cref{tab:proper_training_cityscapes}.
As we mentioned, compared to COCO-style training, we make two different modifications:
\begin{itemize}
    \item Considering the original resolution of the images in Cityscapes dataset \cite{Cityscapes}, which is $2048\times1024$, we replace the standard augmentation of D-DETR designed for COCO by multi-scale training by resizing the shorter side of the image randomly between [$800, 1024$] while limiting the longer side with $2048$ and keeping the aspect ratio. At inference time, we use the original image resolution, that is $2048\times1024$.
    \item Instead of limiting the training of D-DETR by $50$ epochs in COCO-style training, we train them all for $200$ epochs to bring the number of iterations closer between the training regimes for COCO and Cityscapes. While doing that, we keep the learning rate at $2e-4$ with a learning rate drop to $2e-5$ after 160-th epoch.
\end{itemize}

\subsection{Analyses of \gls{LaECE}-style Evaluation}
\gls{LaECE}-style evaluation \cite{saod} relies on \gls{LRP} Error (Eq. \eqref{eq:LRPdefcompact}) and \gls{LaECE} (Eq. \eqref{eq:laece__}) for accuracy and calibration respectively.
Though we are inspried by \cite{saod} for this way of coupling calibration and accuracy, \gls{LaECE}-style evaluation suffers from critical drawbacks on the informativeness of the confidence scores and the dataset design as we discuss below.

\textbf{1. Model-dependent threshold selection.} As \gls{LRP} Error is preferred for accuracy and the thresholds are required to be obtained on the val. set, this way of evaluation satisfies this principle.

\textbf{2. Unambiguous \& fine-grained confidence scores.} Similar to \gls{DECE}, \gls{LaECE} also requires \glspl{FP} to have a confidence of $0.00$ regardless of their localisation quality, which might be useful for the subsequent systems. That is why, as illustrated by the right car in \cref{fig:images}(b), a critical information has been missed once $\tau=0.50$ as its target confidence is set to $0.00$ as suggested in \cite{saod}. 

\textbf{3. Properly-designed datasets.} Another critical drawback of this approach is that the \gls{ID} test set is obtained from a different distribution.
Specifically, the proposed \gls{SAOD} task in \cite{saod} includes two different settings, that are common objects and autonomous vehicle.
In both of the cases, the models are evaluated on a test set collected from a different dataset.
As an example, Obj45K, as a subset of Objects365 \cite{Objects365} dataset is used to evaluate models trained with COCO.
However, as a different dataset introduces domain shift, the settings for \gls{SAOD} task cannot be employed to evaluate the calibration performance for \gls{ID}.
By including the Obj45K split, we demonstrate the effect of domain shifted test set on calibration performance in \cref{tab:coco}.
Specifically, one cannot clearly observe the benefit of post-hoc calibration in \cref{tab:coco} once Obj45K split is used, whereas the post-hoc approaches, which are obtained on \gls{ID} val. set, improve calibration performance of \gls{ID} test set significantly.
This shows that both \gls{ID} and domain-shifted test sets should be part of the evaluation, while this is not the case for \gls{LaECE}-style evaluation.

\textbf{4. Properly-trained detectors \& calibrators.} Finally, this way of evaluation does not have a major issue in terms of the used detectors and calibrators. 
Specifically, four different detectors are used and calibrated with \gls{IR} and \gls{LR} post-hoc approaches.
Among minor issues, one thing to note is that Platt Scaling, as a distribution calibration approach, has not been investigated in \cite{saod}.
Furthermore, the applicability of the calibration approaches are not considered from a broader perspective in terms of detectors.
In this paper, we design \gls{PS} properly, and show that \gls{PS} and \gls{IR} are quite strong baselines in various scenarios including object detection and instance segmentation using very different detectors.

\subsection{Analyses of \gls{CE}-style Evaluation}
\label{sec:CEanalysis}
\gls{CE}-style evaluation thresholds the detectors from $0.50$ to compute  (i) \gls{AP} for accuracy; and (ii) \gls{CE} (Eq. \eqref{eq:ce}) along with \gls{DECE} for calibration.
Another peculiarity of this approach is to employ COCO-style \gls{CE} and \gls{DECE} is the main evaluation measures for calibration performance, which we will provide further details below.

\textbf{1. Model-dependent threshold selection.} As, this type of evaluation also uses a fixed threshold, that is $0.50$, the threshold selection is model-independent. 
Therefore, as the calibration evaluation is quite sensitive to the threshold choice as we showed in \cref{sec:pitfall}, this evaluation approach can also lead to the ambiguity on the best performing detector in terms of calibration. 
In addition, different from \gls{DECE}-style evaluation, the authors also threshold the detection set from $0.50$ while computing the accuracy of the detector using \gls{AP}.
However, \gls{AP} is also quite sensitive to thresholding and can easily mislead the evaluation.
To see that, as an example, please consider \cref{fig:curves}(b) in which we plot \gls{AP} of five different detectors for different confidence score thresholds.
When we use $0.50$, RS R-CNN performs the best, even outperforming Deformable DETR, the most recent detector among all five detectors by around $10$ AP points. 
It also outperforms the recent ATSS by $\sim20$ AP once thresholded from $0.50$.
However, these large gaps in their accuracy only result from the fact that RS R-CNN is more overconfident compared to the other two.
When we use our evaluation approach in \cref{tab:model_zoo_coco}, we can easily see that D-DETR performs $1.4$ LRP better than RS R-CNN and RS R-CNN only outperforms ATSS by $0.8$ LRP, instead of $20$AP.
Therefore, using a fixed threshold on \gls{AP} does not enable the practitioners to compare the accuracy of the detectors.

\textbf{2. Unambiguous \& fine-grained confidence scores.} 
As we briefly discussed in \cref{sec:pitfall}, \cite{popordanoska2024CE} utilizes COCO-style \gls{DECE}, $\mathrm{\text{\gls{DECE}}_C}$, and COCO-style \gls{CE} as the calibration error.
Specifically, $\mathrm{\text{\gls{DECE}}_C}$ (and similarly for \gls{CE}) is the average of 10 $\mathrm{D-ECE}$ values that are obtained for different \gls{IoU} thresholds to validate \glspl{TP}, i.e., from $\tau=0.50$ to $\tau=0.95$ with $0.05$ increments.
However, as we illustrated in the left car of \cref{fig:images}(b), this way of computing $\mathrm{D-ECE}$ can result in the same error value regardless of the estimated confidence score for a detection.
To demonstrate this, 
we start with a simpler version of $\mathrm{\text{\gls{DECE}}_C}$ in which we obtain \gls{DECE} from two \gls{IoU} thresholds as 0.50 and 0.75 ($\mathrm{\text{\gls{DECE}}_{50}}$ and $\mathrm{\text{\gls{DECE}}_{75}}$) and then estimate their average, which can be expressed as:
\begin{align}
   \mathrm{\text{\gls{DECE}}_C}
    &=   \frac{1}{2} (\mathrm{\text{\gls{DECE}}_{50}} + \mathrm{\text{\gls{DECE}}_{75}}) \\ 
    &= \frac{1}{2} \sum_{j=1}^{J} \frac{|\hat{\mathcal{D}}_j|}{|\hat{\mathcal{D}}|} \left\lvert \bar{p}_{j} - \mathrm{precision}_{50}(j)   \right\rvert + \frac{1}{2} \sum_{j=1}^{J} \frac{|\hat{\mathcal{D}}_j|}{|\hat{\mathcal{D}}|} \left\lvert \bar{p}_{j} - \mathrm{precision}_{75}(j)   \right\rvert
\end{align}
where we followed the notation from \cref{sec:relatedwork}, and $\mathrm{precision}_{50}$ and $\mathrm{precision}_{75}$ refer to the precision obtained for $\tau=0.50$ and $\tau=0.75$ respectively.
As we derived in Eq. \eqref{eq:decenormalizedsum} of App. \ref{app:further_details_dece_style}, $\mathrm{\gls{DECE}}$ can be expressed as the normalized sum of bin-wise errors, hence replacing it for each \gls{DECE} with different thresholds:
\begin{align} 
\label{eq:dece_c}
\small
\frac{1}{2|\hat{\mathcal{D}}|} \Bigg(\sum_{j=1}^{J} \Bigg( \Bigg\lvert \sum_{\substack{\hat{b}_i \in \hat{\mathcal{D}}_j \\ \psi_{50}(i) > 0}} \big( \hat{p}_i - 1 \big) + \sum_{\substack{\hat{b}_i \in \hat{\mathcal{D}}_j\\ \psi_{50}(i)=-1}} \hat{p}_i \Bigg\rvert + \Bigg\lvert \sum_{\substack{\hat{b}_i \in \hat{\mathcal{D}}_j \\ \psi_{75}(i) > 0}} \big(\hat{p}_i - 1) + \sum_{\substack{\hat{b}_i \in \hat{\mathcal{D}}_j \\ \psi_{75}(i) = -1}} \hat{p}_i \Bigg\rvert \Bigg) \Bigg)
\end{align}
where $\psi_{50}(i)$ refers to $\psi(i)$ when $\tau=0.50$, and similarly for $0.75$. That is, $\psi_{50}(i)>0$ implies that $i$-th detection is a \gls{TP} for the \gls{IoU} threshold of $\tau=0.50$.
COCO-style \gls{DECE} in Eq. \eqref{eq:dece_c} can yield ambiguous confidence scores for detections with $\psi_{50}(i)>0$ but $\psi_{75}(i)=-1$, that is a detection with \gls{IoU} with the object more than $0.50$ but less than $0.75$.
We now demonstrate this on the example below.

\textbf{Example.} We assume that the detector has a single detection with an \gls{IoU} of $0.60$ and compute the COCO-style \gls{DECE} below by exploiting Eq. \eqref{eq:dece_c}:
\begin{align}
\label{eq:reduced_coco_dece_c}
&\frac{1}{2} \Bigg(\sum_{j=1}^{J} \Bigg( \Bigg\lvert \sum_{\substack{\hat{b}_i \in \hat{\mathcal{D}}_j \\ \psi_{50}(i) > 0}} \big( \hat{p}_i - 1 \big)  \Bigg\rvert + \Bigg\lvert \sum_{\substack{\hat{b}_i \in \hat{\mathcal{D}}_j \\ \psi_{75}(i) = -1}} \hat{p}_i \Bigg\rvert \Bigg) \Bigg) \\
=&\frac{1}{2} \Bigg( \Bigg\lvert \sum_{\substack{\hat{b}_i \in \hat{\mathcal{D}}_j \\ \psi_{50}(i) > 0}} \big( \hat{p}_i - 1 \big)  \Bigg\rvert + \Bigg\lvert \sum_{\substack{\hat{b}_i \in \hat{\mathcal{D}}_j \\ \psi_{75}(i) = -1}} \hat{p}_i \Bigg\rvert \Bigg) \\
=&\frac{1}{2} \Bigg( \Bigg\lvert \hat{p}_i - 1  \Bigg\rvert + \Bigg\lvert \hat{p}_i \Bigg\rvert \Bigg)= \frac{1}{2} \Bigg( 1- \hat{p}_i +\hat{p}_i \Bigg) =0.50
\end{align}

Please note that, Eq. \eqref{eq:reduced_coco_dece_c} shows that COCO-style \gls{DECE} results in a constant value that is independent of the predicted confidence score $\hat{p}_i$.
This simple example can be easily extended to COCO-style \gls{DECE} with 10 different \gls{IoU} thresholds for evaluation resulting in the case in the left car in \cref{fig:images}(b).
As its \gls{IoU} with the object is $0.74$, it will be considered as a \gls{TP} by five \gls{DECE} values with $0.50 \leq \tau <0.75$ and a \gls{FP} for the other five \gls{TP} validation thresholds, i.e. $0.75 \leq \tau <1.00$ in COCO-style \gls{DECE}.
Please note that, this also creates ambiguity while we aim to assign targets to the predictions to obtain post-hoc calibrators as the target confidence of such detections is ambiguous unlike the standard \gls{DECE} or \gls{LaECE}. 
Considering these drawbacks, we assert that COCO-style computation of calibration errors should be avoided.

\textbf{3. Properly-designed datasets.} As we discussed in \cref{sec:intro}, using domain-shifted evaluation sets is crucial for safety-critical applications though they are not used by this way of evaluation.

\textbf{4. Properly-trained detectors \& calibrators.} In terms of baseline calibration methods, \cite{popordanoska2024CE} follows \gls{DECE}-style evaluation and uses \gls{TS} as a baseline.
As we show in this paper that \gls{TS} is not the ideal approach for post-hoc calibration of object detectors, \cite{popordanoska2024CE} also violates this principle.
\renewcommand{\thesection}{C}
\section{Further Details of Our Approach}
\label{app:method}
In this section, we provide further details on our calibration measures, datasets and post-hoc calibrators.

\subsection{Further Details on \gls{LaECE} and \gls{LaACE}}
\label{app:criterion}

In the following we obtain under which condition $\mathrm{LaECE}_0$ and $\mathrm{LaACE}_0$ are minimized.
That is, we show that $\hat{p}_{i}=\mathrm{IoU}(\hat{b}_i, b_{\psi(i)})$ is a sufficient condition for both measures, while it is also necessary condition for $\mathrm{LaACE}_0$.

\textbf{The Optimisation Criterion for $\mathrm{LaECE}_0$}
For class $c$, $\mathrm{LaECE}_0$ is defined as:
\begin{align}
   \mathrm{LaECE}^c_0 
   & = \sum_{j=1}^{J} \frac{|\hat{\mathcal{D}}^{c}_j|}{|\hat{\mathcal{D}}^{c}|} \left\lvert \bar{p}^{c}_{j} - \bar{\mathrm{IoU}}^{c}(j)  \right\rvert,
\end{align}
which can be expressed as,
\begin{align}
 \sum_{j=1}^J \frac{|\hat{\mathcal{D}}_j^c|}{|\hat{\mathcal{D}}^c|} \left\lvert \frac{\sum_{\hat{b}_i \in \hat{\mathcal{D}}_j^c} \hat{p}_{i}}{|\hat{\mathcal{D}}_j^c|} - \frac{\sum_{\hat{b}_i \in \hat{\mathcal{D}}_j^c} \mathrm{IoU}(\hat{b}_i, b_{\psi(i)})}{|\hat{\mathcal{D}}_j^c|}  \right\rvert,
\end{align}
where we replace $\bar{p}_{j}$, the average of the confidence score in bin $j$ by 
$
\bar{p}_{j} = \frac{\sum_{\hat{b}_i \in \hat{\mathcal{D}}_j^c} \hat{p}_{i}}{|\hat{\mathcal{D}}_j^c|}
$
and $ \bar{\mathrm{IoU}}^{c}(j)$ by
\begin{align}
\frac{\sum_{\hat{b}_i \in \hat{\mathcal{D}}_j^c} \mathrm{IoU}(\hat{b}_i, b_{\psi(i)})}{|\hat{\mathcal{D}}_j^c|}.
\end{align}
Cancelling out $|\hat{\mathcal{D}}_j^c|$ as it is a positive number, we have
\begin{align}
\label{eq:laece_alternative}
 \mathrm{LaECE}^c_0 = \frac{1}{|\hat{\mathcal{D}}^c|} \sum_{j=1}^J \left\lvert \sum_{\hat{b}_i \in \hat{\mathcal{D}}_j^c} \hat{p}_{i} - \sum_{\hat{b}_i \in \hat{\mathcal{D}}_j^c} \mathrm{IoU}(\hat{b}_i, b_{\psi(i)})  \right\rvert,
\end{align}
This implies that the calibration error in bin $j$ is minimized once the following expression is minimized
\begin{align}
\label{eq:laece_optimisation_criterion}
\left\lvert \sum_{\hat{b}_i \in \hat{\mathcal{D}}_j^c} \hat{p}_{i} - \sum_{\hat{b}_i \in \hat{\mathcal{D}}_j^c} \mathrm{IoU}(\hat{b}_i, b_{\psi(i)})  \right\rvert=\left\lvert \sum_{\hat{b}_i \in \hat{\mathcal{D}}_j^c} \hat{p}_{i} - \mathrm{IoU}(\hat{b}_i, b_{\psi(i)})  \right\rvert,
\end{align}
implying that $\mathrm{LaECE}_0$ is minimized if $\hat{p}_{i}=\mathrm{IoU}(\hat{b}_i, b_{\psi(i)})$ for all detections.

\textbf{The Optimisation Criterion for $\mathrm{LaACE}_0$} 
For class $c$, $\mathrm{LaACE}_0$ is defined as:
\begin{align}
   \mathrm{LaACE}^c_0 
   =  \sum_{i =1}^{|\hat{\mathcal{D}}^{c}|} \frac{1}{|\hat{\mathcal{D}}^{c}|} \left\lvert \hat{p}_{i} - \mathrm{IoU}(\hat{b}_i, b_{\psi(i)})  \right\rvert,
\end{align}
As $\frac{1}{|\hat{\mathcal{D}}^{c}|}$ is a positive constant, $\mathrm{LaACE}^c_0$ is simply minimized if and only if $\hat{p}_{i} = \mathrm{IoU}(\hat{b}_i, b_{\psi(i)})$ for all detections.

\begin{figure*}[t]
        \centering
            \includegraphics[width=0.6\textwidth]{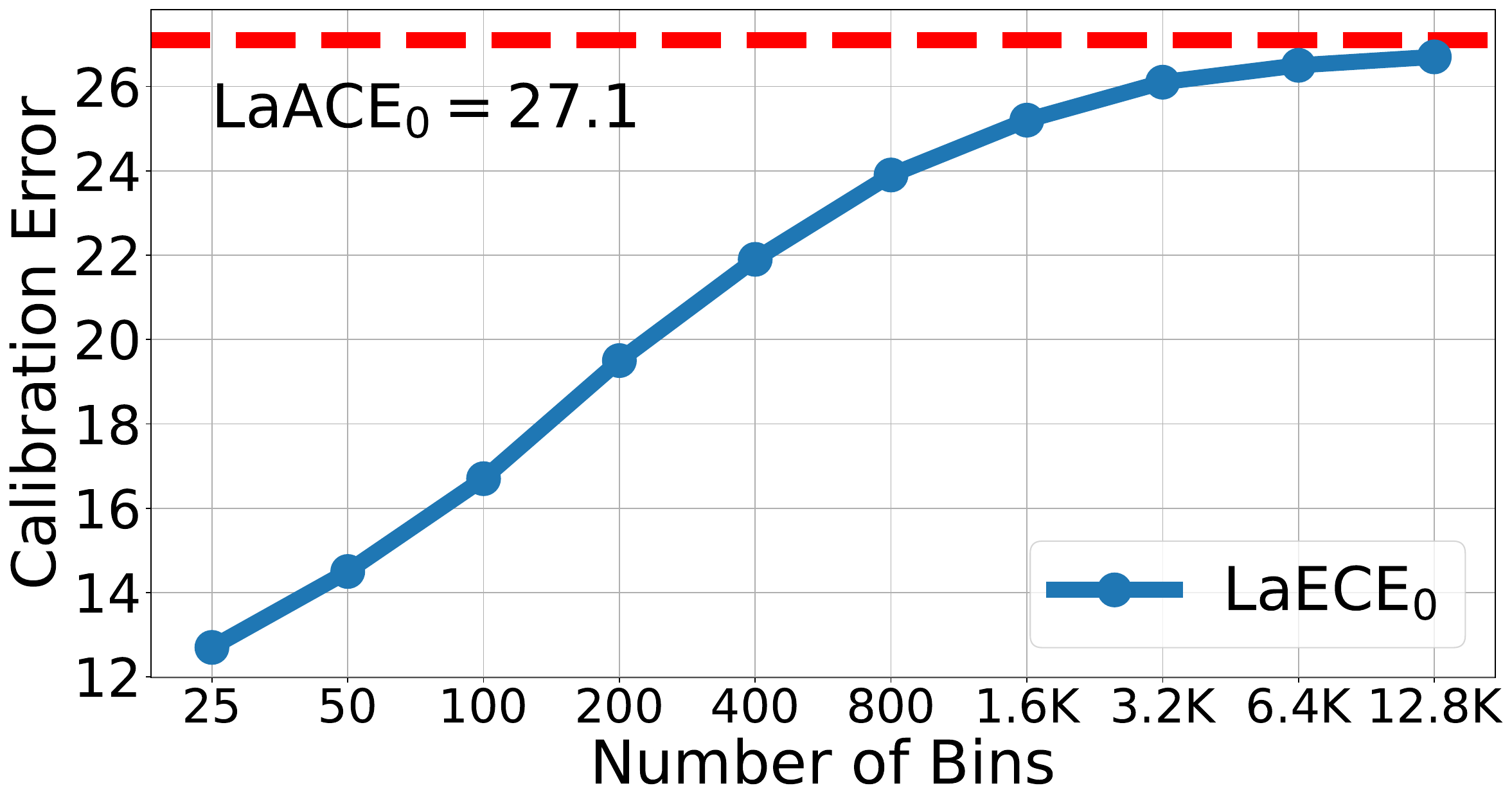}
            \caption{$\mathrm{LaACE}_0$ (\textcolor{red}{red} dashed line at $27.1$) and $\mathrm{LaECE}_0$ over different number of bins (\textcolor{BlueViolet}{blue} curve) using uncalibrated D-DETR on COCO \textit{minitest}.
            The number of bins starts from the original $25$ bins for \gls{LaECE} and gets multiplied up by $2$ for each step.
            $\mathrm{LaECE}_0$ converges to $\mathrm{LaACE}_0$ as the number of bins increases.
            }
        
        \label{fig:app_laece_converges_laace}
\end{figure*}

\textbf{$\mathrm{LaACE}_0 \geq \mathrm{LaECE}_0$ holds.} We now investigate the relationship between  $\mathrm{LaACE}_0$ and $\mathrm{LaECE}_0$, and show that $\mathrm{LaACE}_0$ is greater than or equal to $\mathrm{LaECE}_0$, making $\mathrm{LaACE}_0$ a more challenging measure. 
To show that, we first consider the definition of $\mathrm{LaACE}^c_0$ for class $c$, which is
\begin{align}
   \mathrm{LaACE}^c_0 
   =  \sum_{i =1}^{|\hat{\mathcal{D}}^{c}|} \frac{1}{|\hat{\mathcal{D}}^{c}|} \left\lvert \hat{p}_{i} - \mathrm{IoU}(\hat{b}_i, b_{\psi(i)})  \right\rvert,
\end{align}
which is equal to
\begin{align}
   \mathrm{LaACE}^c_0 
   =  \frac{1}{|\hat{\mathcal{D}}^{c}|} \sum_{j=1}^J  \sum_{\hat{b}_i \in \hat{\mathcal{D}}_j^c} \left\lvert \hat{p}_{i} - \mathrm{IoU}(\hat{b}_i, b_{\psi(i)})  \right\rvert,
\end{align}
as we simply the detections into bins but still compute $\mathrm{LaACE}^c_0$ by measuring the gap between predicted confidence and \gls{IoU} for each detection. Now considering the triangle inequality, which is $|x| + |y| \geq |x + y|$, we can take the absolute value out of the inner summation term,
\begin{align}
   \mathrm{LaACE}^c_0 
   &\geq  \frac{1}{|\hat{\mathcal{D}}^{c}|} \sum_{j=1}^J  \left\lvert \sum_{\hat{b}_i \in \hat{\mathcal{D}}_j^c}  \hat{p}_{i} - \mathrm{IoU}(\hat{b}_i, b_{\psi(i)})  \right\rvert \\
   &=    \frac{1}{|\hat{\mathcal{D}}^c|} \sum_{j=1}^J \left\lvert \sum_{\hat{b}_i \in \hat{\mathcal{D}}_j^c} \hat{p}_{i} - \sum_{\hat{b}_i \in \hat{\mathcal{D}}_j^c} \mathrm{IoU}(\hat{b}_i, b_{\psi(i)})  \right\rvert \\
   &= \mathrm{LaECE}^c_0
\end{align}
Please note that we have already derived the last equality in Eq. \eqref{eq:laece_alternative}, hence enabling us to conclude that $\mathrm{LaACE}_0 \geq \mathrm{LaECE}_0$ holds. Furthermore, considering the extreme case of this inequality, $\mathrm{LaACE}_0 = \mathrm{LaECE}_0$ holds in the case that the number of bins used to compute $\mathrm{LaECE}_0$ is equal to the number of detections. We demonstrate this in \cref{fig:app_laece_converges_laace}, in which the resulting $\mathrm{LaECE}_0$ approximates $\mathrm{LaACE}_0$ as the number of bins used to compute $\mathrm{LaECE}_0$ increases.

\blockcomment{
\subsection{Proof of \cref{theorem:elcerich}}
\label{app:proof}
\begin{theorem}
\label{theorem:elcerich_}
If a detector ensures $\hat{p}_{i}=\mathrm{IoU}(\hat{b}_i, b_{\psi(a)})$ for each detection $i$, implying that $\mathrm{LaECE}_0=\mathrm{LaACE}_0=0$, the confidence scores yielding $\mathrm{\gls{DECE}}=\mathrm{LaECE}_\tau=0$ for any $\tau>0$ can be obtained, but not vice versa. (Please refer to the App. \ref{app:method} for the proof.)
\end{theorem}
\begin{proof}

\end{proof}
}

\textbf{The Cases where There is no Detection From a Class}  We observed that there can be cases where there is no detection for a class.
In such cases, the denominator, $|\hat{\mathcal{D}}^{c}|$, is $0$ in Eq. \eqref{eq:elce} and Eq. \eqref{eq:laace}, making $\mathrm{LaECE}_0$ and $\mathrm{LaACE}_0$ undefined.
To prevent this, we simply ignore such classes while computing the calibration errors.

\subsection{Details of the Datasets Used in our Evaluation Framework}
\label{app:dataset}
In the following, we provide further details on each of the settings that we used in \cref{tab:datasets}.

\textbf{1. Common Objects Setting}
For this setting, we rely on COCO dataset \cite{COCO} which is among the most commonly-used object detection benchmarks.
COCO dataset consists of 80 object classes of varying nature.
As COCO  contains both bounding box and instance mask annotations, we utilise COCO for both of \textit{object detection} and \textit{instance segmentation} in common objects settings.
\paragraph{Training set.} We simply use COCO training split with $118K$ images without any modification.

\paragraph{Validation and \gls{ID} test sets.} As the annotations of the COCO test set are not public, we randomly split the validation set of COCO as minival and minitest sets following the literature \cite{saod, MOCAE}. 
Specifically, both of these sets contain $2.5K$ images, contain objects from each classes in COCO dataset and represent similar characteristics.
As an example, while  COCO minival contains $7.5$ object annotations per image, COCO minitest has $7.2$ object annotations.
\paragraph{Domain-shifted test sets.} With the aim of providing more comprehensive insights regarding the accuracy and the calibration of the detectors, we also evaluate them under certain corruptions.
Specifically, following Oksuz et al. \cite{saod}, we consider $15$ benchmark corruptions from the corruptions provided in \cite{hendrycks2019robustness}.
These corruptions can further be listed as \textit{gaussian noise, shot noise, impulse noise, speckle noise, defocus blur, motion blur, gaussian blur, snow, frost, fog, brightness, contrast, elastic transform, pixelate} and \textit{jpeg compression}.
Furthermore, we only consider each corruption at the severity levels $1$, $2$ and $3$ as it was previously observed that higher severity levels can alter the semantics of the images, especially resulting in some small objects to disappear \cite{saod}.
As a result, we report the average \gls{LRP}, \gls{LaECE} and \gls{LaACE} values over 45 different corruption settings ($15$ corruptions under $3$ severities.), providing a comprehensive evaluation.
Moreover, for more realistic domain shift, we also borrow Obj45K set\cite{saod} which has the same label space with COCO.
Specifically, Obj45K contains $45$K images with $6.0$ object annotations per image.
Even though the label space is the same for both of the COCO minitest and Obj45K datasets, there is still a shift between these datasets as they are obtained from different datasets.
We use Obj45K only for object detection as this dataset does not have mask labels for instance segmentation.
\textbf{2. Autonomous Driving Setting} Cityscapes \cite{Cityscapes} is a well-known autonomous driving dataset consisting of $8$ classes, namely \textit{person}, \textit{rider}, \textit{car}, \textit{truck}, \textit{bus}, \textit{train}, \textit{motorbike} and \textit{bicyle}.
Cityscapes is further used as a common benchmark in the contemporary training-time calibration works for object detection \cite{munir2023caldetr, munir2022tcd, munir2023bpc}.
To provide richer insights across multiple application domains, we also report results on the Cityscapes dataset for \textit{object detection}.
\paragraph{Training set.} We directly use the Cityscapes training split with $2975$ images without any modifications.
 \paragraph{Validation and ID test sets.} As the annotations of the Cityscapes test set are not public either, we randomly split the validation set of Cityscapes as minival and minitest sets following our common objects setting.
 Specifically, both of these sets contain $250$ images, include objects from all of the classes in the Cityscapes dataset and show similar characteristics.
 To exemplify, Cityscapes minival contains $20.7$ object annotations while Cityscapes minitest contains $21.0$ object annotations.
\paragraph{Domain-shifted test sets.} We directly follow the methodology described in COCO minitest-C to construct the Cityscapes minitest-C as the corrupted domain shift set for the autonomous driving setting.
Moreover, for a more realistic domain shift setting, we also consider Foggy Cityscapes \cite{sakaridis2018foggycs} dataset,
consisting of the images in the test set but with an additional realistic fog simulation.
Specifically, there are three different fog severity levels presented in Sakaridis et al. \cite{sakaridis2018foggycs}, one for each of $600$m, $300$m and $150$m meteorological optical ranges (visibility).
During our experiments with Foggy Cityscapes, we select the subset of images that are also present in the Cityscapes minitest dataset to preserve the consistency, yielding a total of $250$ images for each of the three visibility ranges.
We then report the performance by averaging the performance measure over all $3$ visibility ranges.

\textbf{3. Long-tailed Objects Setting} Large Vocabulary Instance Segmentation (LVIS) \cite{LVIS} dataset contains over $1000$ object classes with rich bounding box and instance mask annotations for \textit{object detection} and \textit{instance segmentation} tasks.
Owing to its extremely diverse label set, LVIS comes across as a challenging long-tailed dataset with many rare classes.
For all of our settings, we utilise LVIS v1.0 which builds up on the images of COCO while introducing rich and much more diversified annotations.
\paragraph{Training set.} We directly use the LVIS v1.0 training split with $100K$ samples without any modifications.
\paragraph{Validation and ID test sets.} Similarly with the common objects and autonomous driving settings, we split the LVIS v1.0 validation set into minival and minitest sets.
Specifically, both of these sets contain approximately $9.8$K images and show similar characteristics.
To exemplify, LVIS minival set contains $12.6$ object annotations per image and LVIS minitest set contains $12.4$ object annotations per image.
To enable that post-hoc calibrators are trained properly, we ensure that each class in  LVIS minitest is also represented in LVIS minival while we split LVIS val. set into two.
As some of the classes have very few instances in the val. set (which is only 1 instance for some classes) due to the long-tailed nature of the dataset, this resulted in a case where  LVIS minitest set only includes $935$ classes and  minival set contains  $1035$ classes.
Accordingly, we evaluate the models on those $935$ classes for our long-tailed object setting.
\paragraph{Domain-shifted test sets.} To obtain LVIS minitest-C, we follow our approach used for constructing COCO minitest-C and evaluating it.
\begin{algorithm}[t]
    \caption{Training calibrator on $\valdata$ \label{alg:training}}
    \small
    \begin{algorithmic}[1]
        \Procedure{TrainCalibrator}{$\valdata$}
        \State Cross-validate calibration thr. $\bar{u}^c$ for each class on $\mathcal{D}_{val}$ using LRP with $\tau=0$
        \State Remove detections with score less than $\bar{u}^c$ in $\mathcal{D}_{val}$ to obtain $\mathcal{D}_{thr}$
        \State Train calibrator $\zeta^c(\cdot)$ for each class $c$ on $\mathcal{D}_{thr}$
        \State Calibrate the detections in $\mathcal{D}_{val}$ using $\{\zeta^c(\cdot)\}$ to obtain $\mathcal{D}_{cal}$
        \State Cross-validate operating thr. $\bar{v}^c$ for each class on $\mathcal{D}_{cal}$ using LRP with $\tau$
        \State \textbf{return} $\{\bar{u}^c, \bar{v}^c, \zeta^c(\cdot)\}_{c=1}^{K}$
        \EndProcedure
    \end{algorithmic}
\end{algorithm}

\begin{algorithm}[t]
    \caption{Calibrating detections from an image \label{alg:inference}}
    \small
    \begin{algorithmic}[1]
        \Procedure{Calibrate}{$\{\hat{c}_i, \hat{b}_i, \hat{p}_i\}^N$, $\{\bar{u}^c, \bar{v}^c, \zeta^c(\cdot)\}_{c=1}^{K}$}
        \State Remove detections with score less than $\bar{u}^c$ in $\{\hat{c}_i, \hat{b}_i, \hat{p}_i\}^N$ to obtain $\mathcal{D}_{thr}$
        \State Calibrate confidence scores in $\mathcal{D}_{thr}$, i.e.,  $\hat{p}_i:=\zeta^{\hat{c}_i}(\hat{p}_i)$
        \State Remove detections with calibrated score less than $\bar{v}^c$ in $\mathcal{D}_{thr}$ 
        \State \textbf{return} remaining detections
        \EndProcedure
    \end{algorithmic}
\end{algorithm}

\subsection{Details of Post-hoc Calibrators}
\label{app:posthoc}
This section presents the details of post-hoc calibration methods.

\textbf{Calibrator Training and Inference Algorithms} The details of training and inference with a calibrator are in Alg. \ref{alg:training} and \ref{alg:inference} respectively. In both of the algorithms, we follow the notation that we introduced in \cref{sec:relatedwork} and \cref{sec:posthoc}. Also as an extreme case in which no detection remains after thresholding for a class to train the calibrator ($|\mathcal{D}_{thr}|=0$ in Line 4 of Alg. \ref{alg:training}), we simply use identity function as the calibrator.

\textbf{\gls{NLL} Derivation for Platt Scaling} We now aim to minimize the \gls{NLL} of the predicted calibrated confidence scores ($\hat{p}^{cal}_i$) by considering the target Bernoulli distribution, that is $\mathcal{B}(\mathrm{IoU}(\hat{b}_i, b_{\psi(i)}))$ .
Specifically, using the standard iid assumption, the likelihood of predicted $L$ calibrated probabilities considering the Bernoulli distribution can be expressed as:
\begin{align}
    \prod_{i=1}^{L} \hat{p}_{cal,i}^{{\mathrm{IoU}(b_i, b_{\psi(i)}})} (1-\hat{p}_{cal,i})^{1-\mathrm{IoU}(b_i, b_{\psi(i)})}
\end{align}
where we use $\hat{p}^{cal,i}$ as $\hat{p}^{cal}_i$ for better readability of the notation.
Taking the logarithm and multiplying with $-1$ to make it negative, we have the following expression to minimize:
\begin{align}
    -\sum_{i=1}^{L} \mathrm{IoU}(b_i, b_{\psi(i)}) \log (\hat{p}_{cal,i}) +(1-\mathrm{IoU}(b_i, b_{\psi(i)}) \log(1-\hat{p}_{cal,i}).
\end{align}
Therefore, the \gls{NLL} of the $i$-th example is:
\begin{align}
  - (\mathrm{IoU}(\hat{b}_i, b_{\psi(i)}) \log(\hat{p}_{cal,i}) + (1-\mathrm{IoU}(\hat{b}_i, b_{\psi(i)})) \log(1-\hat{p}_{cal,i})),
\end{align}
which is the cross entropy loss function in Eq. \eqref{eq:nll}.

\renewcommand{\thesection}{D}
\section{Further Experiments}
\label{app:experiments}

This section presents further experiments and analyses that are not included in the paper.

\subsection{Implementation Details}

\textbf{Detectors used for Common Objects Setting} We do not train any detector for this setting and use existing detectors. 
Specifically, for the five training-time calibration methods for object detection in \cref{tab:existing_eval_coco} and \cref{tab:existing_eval_coco}, we borrow the detectors in the official repositories of Cal-DETR and BPC.
Please note that Cal-DETR repository releases all four detectors except BPC.
We also note that among these five approaches, MbLS and MDCA are specifically designed for the classification task, hence their extension to detection are not investigated, and they are used as baselines for TCD, BPC and Cal-DETR.
In the same tables and \cref{tab:ablation}, our baseline D-DETR is taken from mmdetection as we rely on this framework which provides the trained models of several different object detectors.
As for \cref{tab:model_zoo_coco}, we again use mmdetection with the exceptions of: (i) \gls{SOTA} detectors and UP-DETR, which we use their official repositories and (ii) MoCaE which we implement ourselves while fully adhering to the original settings including all of the hyperparameters described in \cite{MOCAE}.
Finally, we again obtain the models mmdetection for instance segmentation task  on \cref{tab:app_coco_instance_segmentation} and \cref{tab:app_coco-c_instance_segmentation}, in which we use a Resnet-50 \cite{ResNet} backbone for all the detectors.

\textbf{Detectors used for Autonomous Driving Setting} For this setting, we train all detectors in \cref{tab:cs} and \cref{tab:existing_eval_cityscapes} as the detectors are not publicly released for this setting.
While doing that we keep all hyperparameters for each detector as it is but only make two changes that we outlined in App. \ref{app:analyses} in the training pipeline to boost the performance of the models and compare them properly.
Specifically, we incorporate this setting into official repositories of Cal-DETR and BPC, and implemented TCD by ourselves as its implementation with D-DETR is not publicly available.
Similarly, please note that, for MbLS and MDCA, two baselines borrowed from classification, their implementation is also not publicly available.
Also considering that extending these methods for object detection requires a thourough thought process and diligent hyperparameter tuning, we do not use these baseline for our autonomous driving setting.
As for D-DETR, we use mmdetection framework.

\textbf{Detectors used for Long-tailed Objects Setting} Similar to the common objects setting, we simply use trained models from mmdetection for this setting.
Specifically, for all three models we used in \cref{tab:lvis}, \cref{tab:app_lvis_platt_scaling_detection} and \cref{tab:app_lvis_platt_scaling_segmentation} for long-tailed detection, we utilise the ones with Resnet-101 backbone.
\begin{table}[t]
\small
\setlength{\tabcolsep}{1.0em}
\centering
\caption{Comparison with \gls{SOTA} methods in terms of other evaluation measures on Cityscapes minitest set. LRP is reported on LRP-optimal thresholds obtained on val. set. AP is reported on top-100 detections. $\tau$ is taken as $0.50$. All measures are lower-better, except AP. \textbf{Bold:} the best, \underline{underlined}: second best. \gls{PS}: Platt Scaling, \gls{IR}: Isotonic Regression. 
}
\label{tab:existing_eval_cityscapes}
\scalebox{1}{
\begin{tabular}{c|c||c|c||c|c||c|c} 
\toprule
\midrule
Cal.&&\multicolumn{2}{c||}{Calibration (thr. 0.30)}&\multicolumn{2}{c||}{Calibration (LRP thr.)}&\multicolumn{2}{c}{Accuracy}\\ 
%
%
Type&Method&\gls{DECE}&$\mathrm{LaECE}$&\gls{DECE}&$\mathrm{LaECE}$&$\mathrm{LRP}$&$\mathrm{AP}\uparrow$ \\ \midrule
Uncal.&D-DETR \cite{DDETR}
&$3.2$&$20.8$
&$3.5$&$20.0$
&$68.4$&$37.3$
\\
\midrule
Training&TCD \cite{munir2022tcd}
&$30.9$&$18.9$
&$31.5$&$18.3$
&$70.5$&$34.2$
\\
Time&BPC \cite{munir2023bpc}
&$8.3$&$26.8$
&$9.2$&$24.9$
&$74.7$&$30.7$
\\
&Cal-DETR \cite{munir2023caldetr}
&$3.7$&$21.4$
&$3.4$&$19.9$
&$68.1$&$37.0$
\\
\midrule
&\gls{PS} for \gls{DECE}
&$\underline{2.8}\imp{0.4}$&$20.2$
&$\underline{2.3}\imp{1.2}$&$19.8$
&$68.4$&$37.3$
\\
Post-hoc&\gls{PS} for \gls{LaECE}
&$14.2$&$\underline{11.3}\imp{7.6}$
&$14.1$&$\mathbf{9.4}\imp{7.9}$
&$68.4$&$37.3$
\\
(Ours)&\gls{IR} for \gls{DECE}
&$\mathbf{1.5}\imp{1.7}$&$19.6$
&$\mathbf{1.4}\imp{2.0}$&$19.4$
&$68.4$&$36.8$
\\
&\gls{IR} for \gls{LaECE}
&$14.2$&$\mathbf{10.4}\imp{8.5}$
&$14.2$&$\mathbf{9.4}\imp{7.9}$
&$68.4$&$36.6$
\\
\midrule
\bottomrule
\end{tabular}}
\end{table}

\subsection{Comparison with SOTA in terms of Other Evaluation Approaches on Autonomous Driving Setting}
In \cref{tab:existing_eval_coco}, we presented that our post-hoc calibrators outperform all existing training-time calibration methods significantly in terms of four different evaluation approaches using existing measures.
Please refer to \ref{sec:experiments} for further details on these evaluation approaches.
We now show that our observations also apply to the autonomous driving dataset.
Specifically, in \cref{tab:existing_eval_coco},  \gls{PS} and \gls{IR} outperform all existing training methods as well as improve the calibration performance of baseline D-DETR on the existing evaluation approaches.

\subsection{Further Experiments with Long-tailed Objects Setting}
In \cref{tab:lvis}, we calibrated the detectors on long-tailed objects setting using \gls{IR}.
To complement that \cref{tab:app_lvis_platt_scaling_detection} and \cref{tab:app_lvis_platt_scaling_segmentation} show the calibration results for \gls{PS}, our other calibrator, demonstrating that \gls{PS} also improves calibration performance by more than $7$ $\mathrm{\gls{LaECE}}_{0}$ and around $2.5$ $\mathrm{\gls{LaACE}}_{0}$.
As a result, \gls{PS} can also be used as a strong baseline on this challenging dataset.

Furthermore, for the sake of completeness, we now evaluate the aforementioned three detectors of the long-tailed setting under domain shift on LVIS minitest-C.
Similarly with \cref{tab:coco} and \cref{tab:cs}, \gls{IR} and \gls{PS} share the top-2 entries on both the object detection (\cref{tab:app_lvis_bbox_corruption}) and the instance segmentation (\cref{tab:app_lvis_segm_corruption}) settings while preserving the accuracy of the models.
As an example, \gls{IR} improves the $\mathrm{\gls{LaECE}}_0$ of the models up to $8.4$ in the object detection and up to $7.6$ in the instance segmentation on LVIS minitest-C.
These results highlight then even under a domain shifted version of a challenging long-tailed dataset, both of \gls{IR} and \gls{PS} remain still quite effective in improving the calibration of the model.

\begin{table}[t]
\small
\setlength{\tabcolsep}{0.2em}
\centering
\caption{Calibrating and evaluating different object detectors for \textit{object detection} on the LVIS minitest dataset. \textbf{Bold:} the best, \underline{underlined}: second best among calibration approaches for each task.}
\label{tab:app_lvis_platt_scaling_detection}
\scalebox{1}{
\begin{tabular}{c|c|c|c||c|c|c||c|c|c} 
\toprule
\midrule
&\multicolumn{3}{c||}{Uncalibrated}&\multicolumn{3}{c||}{Platt Scaling}&\multicolumn{3}{c}{Isotonic Regression}\\
Detector&$\mathrm{\gls{LaECE}}_{0}$&$\mathrm{\gls{LaACE}}_{0}$&$\mathrm{LRP}$&$\mathrm{\gls{LaECE}}_{0}$&$\mathrm{\gls{LaECE}}_{0}$&$\mathrm{LRP}$&$\mathrm{\gls{LaECE}}_{0}$&$\mathrm{\gls{LaECE}}_{0}$&$\mathrm{LRP}$ \\ \midrule
Mask R-CNN \cite{MaskRCNN}
&$25.2$&$30.4$&$74.7$
&$\underline{18.2}$&$\underline{28.4}$&$74.7$
&$\mathbf{17.1}$&$\mathbf{28.0}$&$74.6$
\\
Seesaw Mask R-CNN \cite{seesawloss}
&$25.0$&$30.2$&$73.1$
&$\underline{18.4}$&$\underline{28.3}$&$73.1$
&$\mathbf{16.8}$&$\mathbf{27.8}$&$73.0$
\\
Seesaw Cascade R-CNN \cite{seesawloss}
&$26.4$&$31.5$&$70.7$
&$\underline{19.0}$&$\underline{28.8}$&$70.7$
&$\mathbf{17.4}$&$\mathbf{28.7}$&$70.8$
\\
\midrule
\bottomrule
\end{tabular}}
\end{table}
\begin{table}[t]
\small
\setlength{\tabcolsep}{0.2em}
\centering
\caption{Calibrating and evaluating different object detectors for \textit{instance segmentation} on the LVIS minitest dataset. \textbf{Bold:} the best, \underline{underlined}: second best among calibration approaches for each task.}
\label{tab:app_lvis_platt_scaling_segmentation}
\scalebox{0.95}{
\begin{tabular}{c|c|c|c||c|c|c||c|c|c} 
\toprule
\midrule
&\multicolumn{3}{c||}{Uncalibrated}&\multicolumn{3}{c||}{Platt Scaling}&\multicolumn{3}{c}{Isotonic Regression}\\
Detector&$\mathrm{\gls{LaECE}}_{0}$&$\mathrm{\gls{LaACE}}_{0}$&$\mathrm{LRP}$&$\mathrm{\gls{LaECE}}_{0}$&$\mathrm{\gls{LaECE}}_{0}$&$\mathrm{LRP}$&$\mathrm{\gls{LaECE}}_{0}$&$\mathrm{\gls{LaECE}}_{0}$&$\mathrm{LRP}$ \\ \midrule
Mask R-CNN \cite{MaskRCNN}
&$25.5$&$30.6$&$75.3$
&$\underline{18.6}$&$\underline{28.6}$&$75.3$
&$\mathbf{17.6}$&$\mathbf{28.4}$&$75.3$
\\
Seesaw Mask R-CNN \cite{seesawloss}
&$25.0$&$30.2$&$73.7$
&$\underline{18.0}$&$\underline{27.8}$&$73.7$
&$\mathbf{16.7}$&$\mathbf{27.6}$&$73.7$
\\
Seesaw Cascade R-CNN \cite{seesawloss}
&$25.5$&$30.6$&$71.7$
&$\underline{18.4}$&$\underline{28.2}$&$71.7$
&$\mathbf{17.2}$&$\mathbf{28.1}$&$71.6$
\\
\midrule
\bottomrule
\end{tabular}}
\end{table}

\begin{table}[t]
\small
\setlength{\tabcolsep}{0.2em}
\centering
\caption{Calibrating and evaluating different object detectors for \textit{object detection} on the LVIS minitest-C domain shift dataset. \textbf{Bold:} the best, \underline{underlined}: second best among calibration approaches for each task.}
\label{tab:app_lvis_bbox_corruption}
\scalebox{1}{
\begin{tabular}{c|c|c|c||c|c|c||c|c|c} 
\toprule
\midrule
&\multicolumn{3}{c||}{Uncalibrated}&\multicolumn{3}{c||}{Platt Scaling}&\multicolumn{3}{c}{Isotonic Regression}\\
Detector&$\mathrm{\gls{LaECE}}_{0}$&$\mathrm{\gls{LaACE}}_{0}$&$\mathrm{LRP}$&$\mathrm{\gls{LaECE}}_{0}$&$\mathrm{\gls{LaECE}}_{0}$&$\mathrm{LRP}$&$\mathrm{\gls{LaECE}}_{0}$&$\mathrm{\gls{LaECE}}_{0}$&$\mathrm{LRP}$ \\ \midrule
Mask R-CNN \cite{MaskRCNN}
&$25.9$&$30.7$&$83.8$
&$\underline{19.8}$&$\mathbf{28.6}$&$83.8$
&$\mathbf{18.9}$&$\underline{28.8}$&$83.8$
\\
Seesaw Mask R-CNN \cite{seesawloss}
&$26.3$&$30.7$&$83.3$
&$\underline{20.0}$&$\mathbf{28.3}$&$83.3$
&$\mathbf{19.0}$&$\underline{28.6}$&$83.3$
\\
Seesaw Cascade R-CNN \cite{seesawloss}
&$27.9$&$32.3$&$82.0$
&$\underline{20.5}$&$\mathbf{28.9}$&$82.0$
&$\mathbf{19.5}$&$\underline{29.3}$&$82.0$
\\
\midrule
\bottomrule
\end{tabular}}
\end{table}

\begin{table}[t]
\small
\setlength{\tabcolsep}{0.2em}
\centering
\caption{Calibrating and evaluating different object detectors for \textit{instance segmentation} on the LVIS minitest-C domain shift dataset. \textbf{Bold:} the best, \underline{underlined}: second best among calibration approaches for each task.}
\label{tab:app_lvis_segm_corruption}
\scalebox{1}{
\begin{tabular}{c|c|c|c||c|c|c||c|c|c} 
\toprule
\midrule
&\multicolumn{3}{c||}{Uncalibrated}&\multicolumn{3}{c||}{Platt Scaling}&\multicolumn{3}{c}{Isotonic Regression}\\
Detector&$\mathrm{\gls{LaECE}}_{0}$&$\mathrm{\gls{LaACE}}_{0}$&$\mathrm{LRP}$&$\mathrm{\gls{LaECE}}_{0}$&$\mathrm{\gls{LaECE}}_{0}$&$\mathrm{LRP}$&$\mathrm{\gls{LaECE}}_{0}$&$\mathrm{\gls{LaECE}}_{0}$&$\mathrm{LRP}$ \\ \midrule
Mask R-CNN \cite{MaskRCNN}
&$25.6$&$30.3$&$84.4$
&$\underline{19.4}$&$\mathbf{28.1}$&$84.4$
&$\mathbf{18.7}$&$\underline{28.4}$&$84.4$
\\
Seesaw Mask R-CNN \cite{seesawloss}
&$25.7$&$30.1$&$83.9$
&$\underline{19.7}$&$\mathbf{27.9}$&$83.9$
&$\mathbf{18.8}$&$\underline{28.4}$&$83.9$
\\
Seesaw Cascade R-CNN \cite{seesawloss}
&$26.6$&$30.9$&$82.8$
&$\underline{19.8}$&$\mathbf{28.1}$&$82.8$
&$\mathbf{19.0}$&$\underline{28.5}$&$82.8$
\\
\midrule
\bottomrule
\end{tabular}}
\end{table}

\subsection{Instance Segmentation on Common Objects Setting} 
In addition to evaluating the object detectors with common objects, we now evaluate instance segmentation models.
To calibrate these models, we use mask \gls{IoU} as the target for our calibration measures.
For our experiments in this setting, we utilise three well-known detectors, namely HTC \cite{HTC} , Queryinst \cite{queryinst}, and Mask2Former \cite{cheng2021mask2former}.
\cref{tab:app_coco_instance_segmentation} presents the results on COCO minitest, where we can observe that our \gls{IR} improves the $\mathrm{\gls{LaECE}}_0$ of the models significantly by up to $23.8$ and $\mathrm{\gls{LaACE}}_0$ by up to $9.9$.
\gls{PS} generally perform similar to \gls{IR} in terms of $\mathrm{\gls{LaACE}}_0$ but slightly worse on $\mathrm{\gls{LaECE}}_0$.
Analogously with the ID test set, we observe drastic calibration improvements with our \gls{IR} for domain-shifted test set in \cref{tab:app_coco-c_instance_segmentation}, in which it improve $\mathrm{\gls{LaECE}}_0$ up to $22.6$.
These results further validate the effectiveness of our \gls{IR} and \gls{PS} on instance segmentation task.

\begin{table}[t]
\small
\setlength{\tabcolsep}{0.2em}
\centering
\caption{Calibrating and evaluating different object detectors for \textit{instance segmentation}. We use our common objects setting and report the results on COCO \textit{minitest}. \textbf{Bold:} the best, \underline{underlined}: second best among calibration approaches for each task.}
\label{tab:app_coco_instance_segmentation}
\scalebox{1.00}{
\begin{tabular}{c|c|c|c||c|c|c||c|c|c} 
\toprule
\midrule
&\multicolumn{3}{c||}{Uncalibrated}&\multicolumn{3}{c||}{Platt Scaling}&\multicolumn{3}{c}{Isotonic Regression}\\
Detector&$\mathrm{\gls{LaECE}}_{0}$&$\mathrm{\gls{LaACE}}_{0}$&$\mathrm{LRP}$&$\mathrm{\gls{LaECE}}_{0}$&$\mathrm{\gls{LaECE}}_{0}$&$\mathrm{LRP}$&$\mathrm{\gls{LaECE}}_{0}$&$\mathrm{\gls{LaECE}}_{0}$&$\mathrm{LRP}$ \\ \midrule
HTC \cite{HTC}
&$26.2$&$29.1$&$60.5$
&$\underline{10.2}$&$\underline{23.2}$&$60.5$
&$\mathbf{7.8}$&$\mathbf{22.3}$&$60.5$
\\
QueryInst \cite{queryinst}
&$11.5$&$23.8$&$56.4$
&$\underline{10.0}$&$\underline{22.8}$&$56.4$
&$\mathbf{8.2}$&$\mathbf{21.9}$&$56.4$
\\
Mask2Former \cite{cheng2021mask2former}
&$31.3$&$32.1$&$54.1$
&$\underline{9.6}$&$\underline{22.4}$&$54.1$
&$\mathbf{7.5}$&$\mathbf{22.2}$&$54.2$
\\
\midrule
\bottomrule
\end{tabular}}
\end{table}
\begin{table}[t]
\small
\setlength{\tabcolsep}{0.2em}
\centering
\caption{Calibrating and evaluating different object detectors for \textit{instance segmentation} under domain shift. We use our common objects setting and report the results on COCO minitest-C domain shift dataset. \textbf{Bold:} the best, \underline{underlined}: second best among calibration approaches for each task.}
\label{tab:app_coco-c_instance_segmentation}
\scalebox{1.00}{
\begin{tabular}{c|c|c|c||c|c|c||c|c|c} 
\toprule
\midrule
&\multicolumn{3}{c||}{Uncalibrated}&\multicolumn{3}{c||}{Platt Scaling}&\multicolumn{3}{c}{Isotonic Regression}\\
Detector&$\mathrm{\gls{LaECE}}_{0}$&$\mathrm{\gls{LaACE}}_{0}$&$\mathrm{LRP}$&$\mathrm{\gls{LaECE}}_{0}$&$\mathrm{\gls{LaECE}}_{0}$&$\mathrm{LRP}$&$\mathrm{\gls{LaECE}}_{0}$&$\mathrm{\gls{LaECE}}_{0}$&$\mathrm{LRP}$ \\ \midrule
HTC \cite{HTC}
&$26.8$&$30.0$&$73.9$
&$\underline{12.6}$&$\underline{24.8}$&$73.9$
&$\mathbf{10.5}$&$\mathbf{24.4}$&$73.9$
\\
QueryInst \cite{queryinst}
&$13.0$&$25.2$&$69.9$
&$\underline{11.8}$&$\underline{24.2}$&$69.9$
&$\mathbf{9.8}$&$\mathbf{23.5}$&$70.0$
\\
Mask2Former \cite{cheng2021mask2former}
&$32.8$&$33.4$&$67.8$
&$\underline{12.3}$&$\mathbf{24.1}$&$67.8$
&$\mathbf{10.2}$&$\underline{24.3}$&$67.8$
\\
\midrule
\bottomrule
\end{tabular}}
\end{table}

\begin{table}[t]
\small
\setlength{\tabcolsep}{0.1em}
\centering
\caption{Ablation experiments on Isotonic Regression using D-DETR. \textbf{Bold:} the best, \underline{underlined}: second best. \xmark \  denotes that a domain-shifted val. set is used for obtaining thresholds and calibration, resulting in a big drop in accuracy (\textcolor{red}{red} font).}
\label{tab:app_ablation_isotonic_regression}
\scalebox{1.00}{
\begin{tabular}{c||c|c||c||c|c|c||c|c|c} 
\toprule
\midrule
&\multicolumn{2}{c||}{Ablations on Dataset}&\multicolumn{1}{c||}{Ablations on Model} & \multicolumn{3}{c||}{COCO \textit{minitest}} & \multicolumn{3}{c}{Cityscapes \textit{minitest}}\\%

Method&ID Val. Set&Threshold&Class-wise&$\mathrm{LaECE}_{0}$&$\mathrm{LaACE}_{0}$&$\mathrm{LRP}$ &$\mathrm{LaECE}_{0}$&$\mathrm{LaACE}_{0}$&$\mathrm{LRP}$ \\ \midrule
&\xmark& &
&$14.6$& $25.1$ &\textcolor{red}{$61.2$} 
& $10.5$&$\mathbf{23.0}$&\textcolor{red}{$60.2$} \\
\multirow{3}{*}{Isotonic Regression}
&\cmark& &
&$10.3$& $23.6$ & $57.1$ 
& $12.1$ &$25.8$&$57.5$ \\
&\cmark&\cmark&
&$9.8$& $24.0$ & $57.2$ 
&$11.3$&$26.1$&$57.2$ \\
&\cmark& &\cmark 
&$\textbf{7.5}$& $23.2$&$58.0$ 
&$\mathbf{8.6}$ &$23.8$&$56.2$ \\
&\cmark&\cmark&\cmark
&$\underline{7.7}$& $\textbf{23.1}$ & $57.2$ 
&$\underline{9.0}$&$\underline{23.7}$&$56.8$ \\
\bottomrule
\end{tabular}}
\end{table}

\begin{table}[t]
\small
\setlength{\tabcolsep}{0.1em}
\centering
\caption{Ablation experiments on post-hoc calibrators using D-DETR. \textbf{Bold:} the best, \underline{underlined}: second best. \xmark \  denotes that a domain-shifted val. set is used for obtaining thresholds and calibration, resulting in a big drop in accuracy (\textcolor{red}{red} font). Bias term only exists in the formulation of \gls{PS} ($b$ in Eq. \eqref{eq:ps}), hence it is N/A for \gls{IR}.}
\label{tab:app_ablation_dece}
\scalebox{1}{
\begin{tabular}{c||c|c||c||c|c||c|c} 
\toprule
\midrule
&\multicolumn{2}{c||}{Ablations on Dataset}& Model & \multicolumn{2}{c||}{COCO \textit{minitest}} & \multicolumn{2}{c}{Cityscapes \textit{minitest}}\\%

Method&ID Val. Set&Threshold&Bias Term&$\text{\gls{DECE}}$&$\mathrm{LRP}$ &$\text{\gls{DECE}}$&$\mathrm{LRP}$ \\ \midrule
&\xmark& &
&\underline{$8.0$}&\textcolor{red}{$69.8$}
&$\mathbf{2.7}$&\textcolor{red}{$71.7$} \\
Platt&\cmark& &
&$10.0$&$66.3$
&$4.3$&$68.4$ \\
Scaling
&\cmark&\cmark&
&$10.0$&$66.3$
&$3.6$&$68.4$ \\
&\cmark&\cmark&\cmark
&$\mathbf{2.4}$&$66.3$
&$\underline{3.2}$&$68.4$ \\

\midrule
&\xmark& &N/A
&$13.2$&\textcolor{red}{$69.4$}
&$6.1$&\textcolor{red}{$71.9$}\\
Isotonic&\cmark& &N/A
&$\mathbf{1.5}$&$66.0$
&$\underline{0.8}$&$68.7$ \\
Regression&\cmark&\cmark&N/A
&\underline{$2.6$}&$66.2$
&$\mathbf{0.4}$&$68.4$ \\
\bottomrule
\end{tabular}}
\end{table}

\subsection{Further Analyses}

\textbf{Further Ablations} Similar to \cref{tab:ablation}, we perform ablations over different design choices for \gls{IR} in \cref{tab:app_ablation_isotonic_regression} using both COCO-minitest and Cityscapes-minitest.
As is the case with \gls{TS}, domain-shifted val. set degrades the accuracy of the detector, in red font, as the operating thresholds obtained on these val. sets do not generalize to the \gls{ID} test set.
Our final design with thresholding and class-wise calibrators reaches the best or second best performance in terms of all calibration measures, validating our design choice on post-hoc calibrators.

Furthermore, we analyse the behavior of our \gls{PS} and \gls{IR} under different design choices for \gls{DECE} as a different calibration measure in \cref{tab:app_ablation_dece}.
We note that, as a fixed threshold is not a good approach for evaluation, here we compute \gls{DECE} using LRP-optimal thresholding similar to computing our calibration measures.
Accordingly, as discussed in \cref{sec:posthoc}, we construct target and prediction pairs to train calibrators by considering \gls{DECE} instead of our localisation-based calibration measures.
\cref{tab:app_ablation_dece} shows that our design choices also generalize to \gls{DECE} as using \gls{ID} val. set, thresholding the detections and using a bias term either perform the best or the second best in terms of \gls{DECE} also by preserving the accuracy of the detector.
As an example, bias term significantly helps for calibration for COCO minitest, reaching $2.4$ \gls{DECE} decreasing it from $10.0$ compared to not using bias term.
These results also validates our design choices.

\begin{figure*}[t]
        \captionsetup[subfigure]{}
        \centering
        \begin{subfigure}[b]{0.32\textwidth}
        \centering
            \includegraphics[width=\textwidth]{up-detr_uncalibrated_reliability_diagram.pdf}
            \caption{Uncalibrated UP-DETR}
        \end{subfigure}
        \begin{subfigure}[b]{0.32\textwidth}
        \centering
            \includegraphics[width=\textwidth]{up-detr_isotonic_regression_reliability_diagram.pdf}
            \caption{UP-DETR with IR}
        \end{subfigure}
        \begin{subfigure}[b]{0.32\textwidth}
        \centering
            \includegraphics[width=\textwidth]{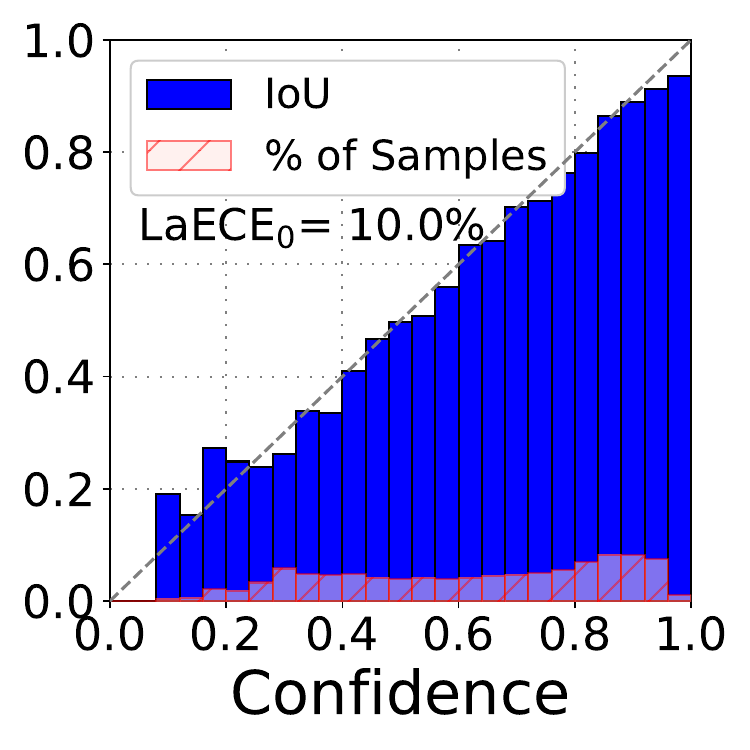}
            \caption{UP-DETR with PS }
        \end{subfigure}
        \caption{Uncalibrated UP-DETR \cite{dai2022updetr} (a), calibrated UP-DETR \cite{dai2022updetr} with isotonic regression (b) and calibrated UP-DETR \cite{dai2022updetr} with platt scaling (c) reliability diagrams on COCO \textit{minitest}\cite{COCO}.
        \label{fig:app_updetr_coco_reliability}}
\end{figure*}

\begin{figure*}[t]
        \captionsetup[subfigure]{}
        \centering
        \begin{subfigure}[b]{0.32\textwidth}
        \centering
            \includegraphics[width=\textwidth]{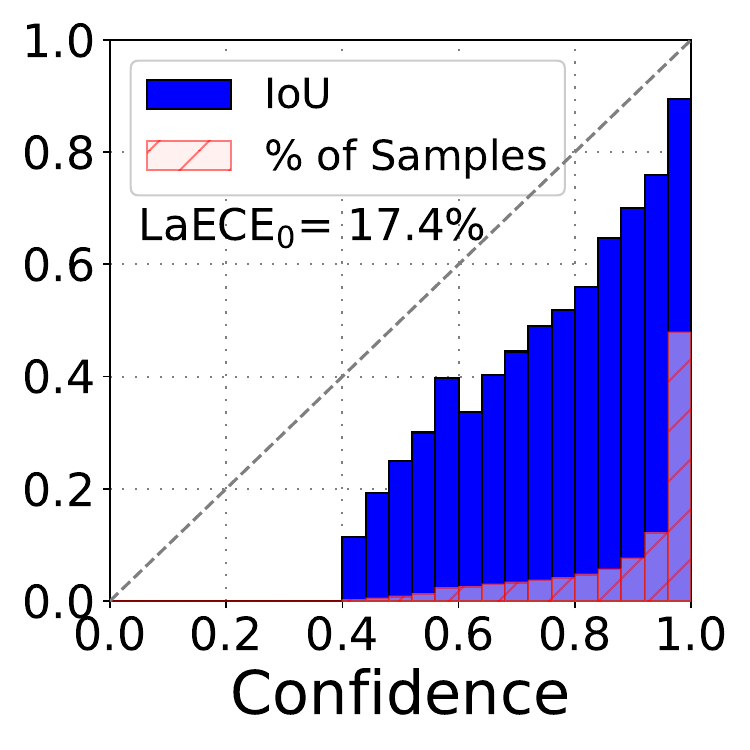}
            \caption{Uncalibrated EVA}
        \end{subfigure}
        \begin{subfigure}[b]{0.32\textwidth}
        \centering
            \includegraphics[width=\textwidth]{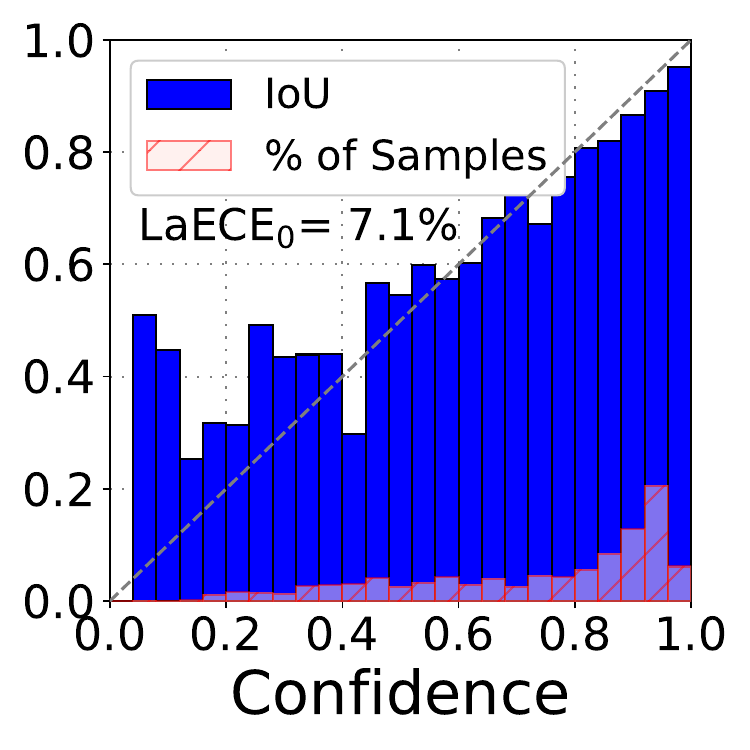}
            \caption{EVA with IR}
        \end{subfigure}
        \begin{subfigure}[b]{0.32\textwidth}
        \centering
            \includegraphics[width=\textwidth]{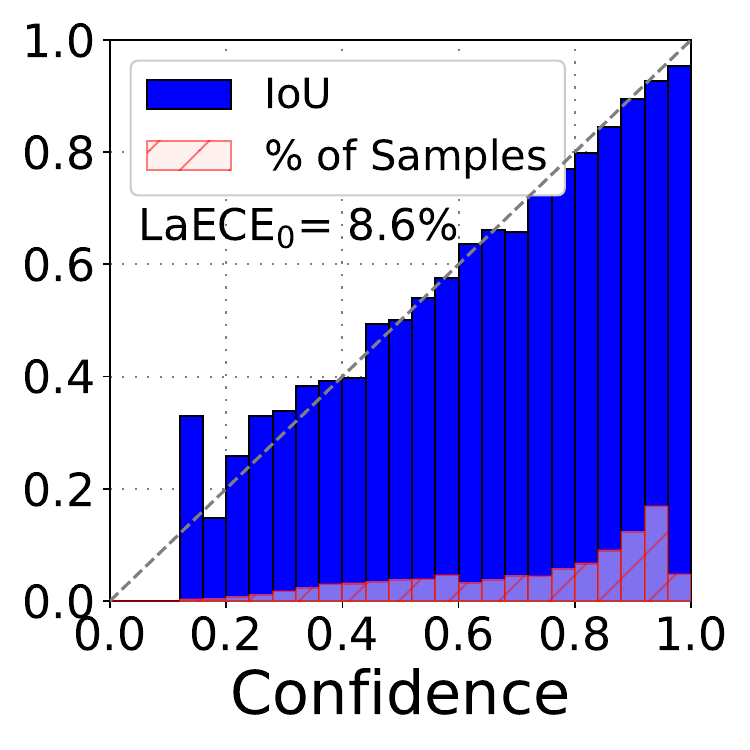}
            \caption{EVA with PS }
        \end{subfigure}

        \caption{Uncalibrated EVA \cite{EVA} (a), calibrated EVA \cite{EVA} with isotonic regression (b) and calibrated EVA \cite{EVA} with platt scaling (c) reliability diagrams on COCO \textit{minitest}\cite{COCO}.
        }
        \label{fig:eva_coco_reliability}
\end{figure*}

\begin{figure*}[t]
        \captionsetup[subfigure]{}
        \centering
        \begin{subfigure}[b]{0.32\textwidth}
        \centering
            \includegraphics[width=\textwidth]{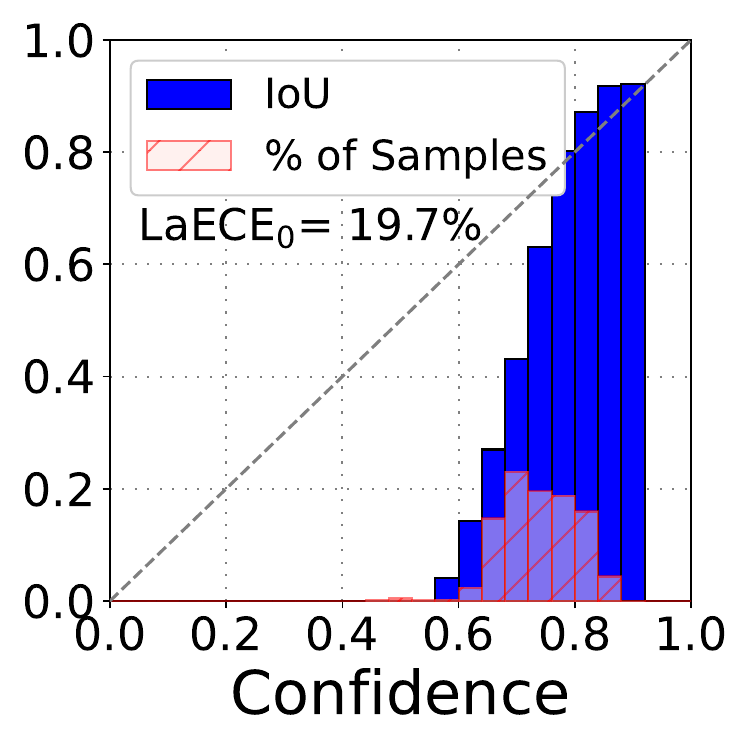}
            \caption{Uncalibrated RS R-CNN}
        \end{subfigure}
        \begin{subfigure}[b]{0.32\textwidth}
        \centering
            \includegraphics[width=\textwidth]{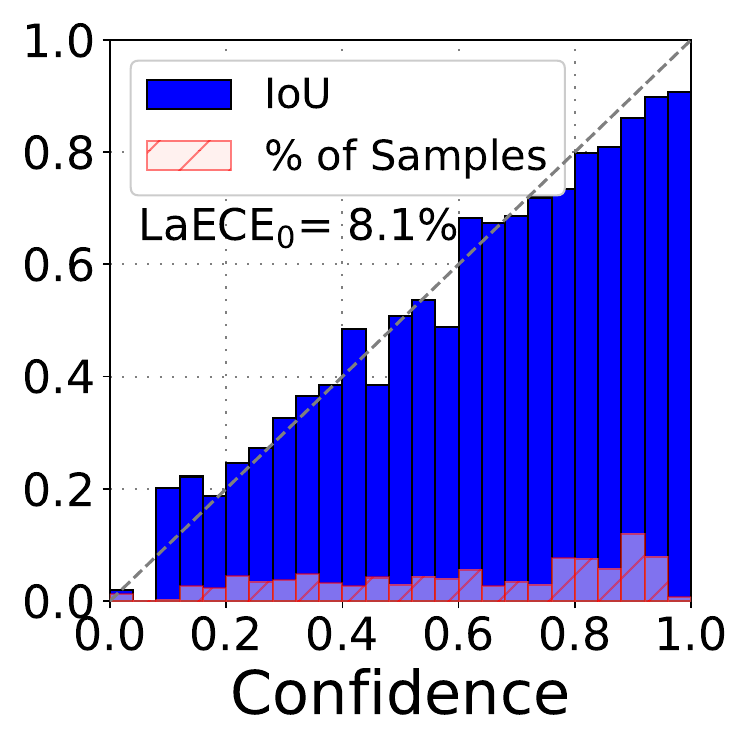}
            \caption{RS R-CNN with IR}
        \end{subfigure}
        \begin{subfigure}[b]{0.32\textwidth}
        \centering
            \includegraphics[width=\textwidth]{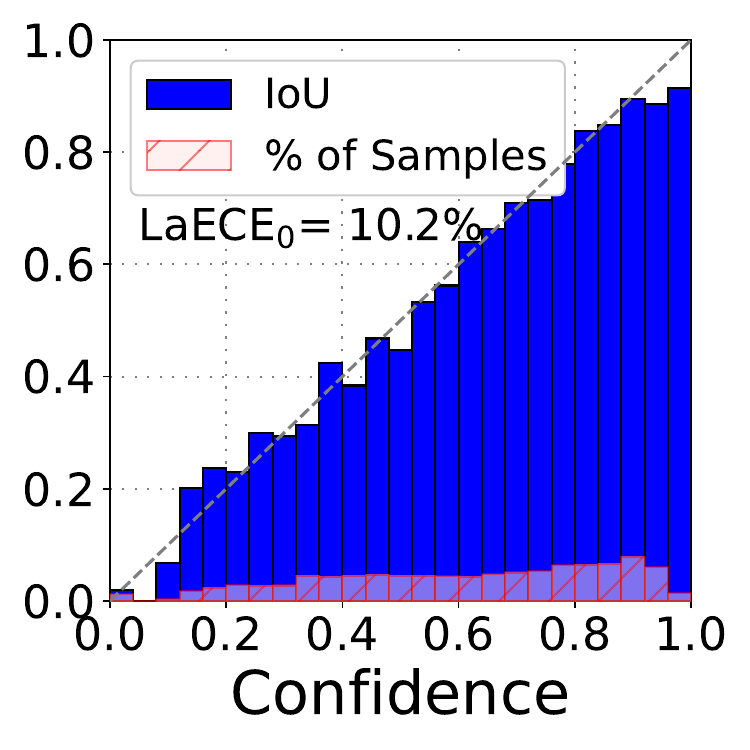}
            \caption{RS R-CNN with PS }
        \end{subfigure}

        \caption{Uncalibrated RS R-CNN \cite{RSLoss} (a), calibrated RS R-CNN \cite{RSLoss} with isotonic regression (b) and calibrated RS R-CNN \cite{RSLoss} with platt scaling (c) reliability diagrams on COCO \textit{minitest}\cite{COCO}.
        }
        \label{fig:rs_rcnn_coco_reliability}
\end{figure*}

\textbf{More Reliability Diagrams} In this part, we further provide reliability diagrams for three models from \cref{tab:model_zoo_coco}, namely: (i) UP-DETR in \cref{fig:app_updetr_coco_reliability}; (ii) EVA \cite{EVA} in \cref{fig:eva_coco_reliability}; and (iii) RS R-CNN in \cref{fig:rs_rcnn_coco_reliability}.
The improvements provided by our PS and IR are evident in the reliability diagrams as well in line with the results of \cref{tab:model_zoo_coco}.


\textbf{Comparing the Detectors in \cref{fig:question}} 
We used five uncalibrated detectors (marked with $^*$ in \cref{tab:model_zoo_coco}) in \cref{fig:question} to illustrate how challenging evaluating the object detectors are. 
We now compare these detectors using our evaluation framework.
\cref{tab:model_zoo_coco} shows that D-DETR performs the best whereas Faster R-CNN performs the worst in terms of both accuracy ($57.3$ vs. $60.4$ \gls{LRP}) and calibration ($12.7$ vs. $27.0$ $\mathrm{\gls{LaECE}}_{0}$).
This is an expected result as (i) D-DETR and Faster R-CNN are trained with Focal Loss \cite{FocalLoss} and Cross-entropy loss, between which Focal Loss provides better calibration \cite{FocalLoss_Calibration}; and (ii) D-DETR is more accurate than Faster R-CNN \cite{DDETR}.

\titlespacing{\paragraph}{%
 5pt}{%
 5pt}{0.5em}%

\end{document}